\documentclass[journal,12pt,onecolumn,draftclsnofoot]{IEEEtran}

\ifCLASSINFOpdf
\else
\fi
\usepackage{lineno,hyperref}
\usepackage{array}

\usepackage{amsmath}
\usepackage{amssymb}  
\usepackage{hyperref}       
\usepackage{url}            
\usepackage{graphicx}
\usepackage{comment}
\usepackage{enumitem}

\newtheorem{theorem}{Theorem}

\newtheorem{lemma}{Lemma}
\newtheorem{corollary}{Corollary}
\newtheorem{remark}{Remark}

\newcommand{\range}{\operatorname{range}}
\newcommand{\dom}{\operatorname{dom}}
\newcommand{\Var}{\operatorname{Var}}
\newcommand{\cov}{\operatorname{Cov}}
\newcommand{\req}[1]{Eq.\,(\ref{#1})}

\begin{document}
%
\title{Optimizing Variational Representations of Divergences 
and Accelerating their Statistical Estimation}
%
%
%

\author{Jeremiah~Birrell, Markos~A.~Katsoulakis and Yannis~Pantazis
\thanks{ J. Birrell and M.A. Katsoulakis  are with the Department of Mathematics and Statistics, University of Massachusetts Amherst, Amherst, MA 01003,  USA (email: birrell@math.umass.edu, markos@math.umass.edu).}
\thanks{Y. Pantazis is with  the Institute of Applied and Computational Mathematics, Foundation for Research and Technology, 
   Hellas, Heraklion, GR-70013, Greece (email: pantazis@iacm.forth.gr).}
}

%
%



\maketitle

\begin{abstract}
Variational representations of divergences and distances between high-dimensional probability distributions offer significant theoretical insights and practical advantages in numerous  research areas. Recently, they have gained  popularity in machine learning  as a tractable and scalable  approach for training probabilistic models and for statistically differentiating between data distributions. Their advantages include: 1) They can be estimated from data as statistical averages. 2) Such representations can leverage  the ability of neural networks to  efficiently approximate optimal solutions in   function spaces.  However, a systematic and practical approach  to  improving the tightness of such variational formulas, and accordingly accelerate statistical  learning and estimation from data, is currently lacking. Here we develop such a methodology for building new, tighter  variational representations of divergences. Our approach relies on  improved objective functionals constructed via an    auxiliary  optimization  problem. Furthermore, the calculation of the functional Hessian of objective functionals unveils the local curvature differences around the common optimal variational solution; this quantifies and orders the tightness gains between different variational representations. Finally, numerical simulations utilizing neural network optimization demonstrate  that tighter representations can result in significantly faster  learning and more accurate estimation of divergences in both synthetic and real datasets (of more than 1000 dimensions),  often accelerated  by nearly an order of magnitude.

\end{abstract}
\newpage


%
\IEEEpeerreviewmaketitle

\section{Introduction}
%
%
%
%
Divergences and distances between multivariate probability distributions play a central role in many mathematical, engineering, and scientific fields ranging from statistical physics, large deviations theory, and uncertainty quantification to information theory, statistics, and machine learning. Variational representation formulas for divergences, also referred to as dual formulations, convert  divergence calculation into an optimization problem over a function space and   offer a valuable mathematical tool to build, train, and analyze probabilistic models and measure the similarity between data collections. Typical examples of variational representations are, among others, the Legendre transformation (LT) of an $f$-divergence \cite{Renyi1961,Ali1966}, the Donsker-Varadhan (DV) formula for the Kullback-Leibler (KL) divergence  \cite{DV1983, Dupuis_Ellis} and the Rubinstein-Kantorovich duality formula for Wasserstein distance \cite{villani2008optimal}. 
Variational representations have been used in statistical mechanics and interacting particles systems  \cite{Kipnis:99}, large deviations \cite{Dupuis_Ellis}, divergence estimation \cite{Nguyen_2007, Nguyen_Full_2010, Ruderman}, determining variable independence through mutual information estimation \cite{MINE_paper}, adversarial learning of generative models \cite{GAN, f_GAN, WGAN}, uncertainty quantification (UQ) of stochastic processes \cite{chowdhary_dupuis_2013,DKPP}, bounding risk  in  probably approximately correct (PAC) learning \cite{10.1145/307400.307435,10.1145/267460.267466,catoni2008pac}, 
as well as in parameter estimation \cite{Broniatowski_Keziou}. 

Two main mathematical ingredients are involved in the construction of a variational formula. 
First, the {\it function space} where the optimal solution will be searched for and, second, the representation expression, called here the `{\it objective functional}', whose optimization leads to the value of the divergence. Crucial practical advantages of variational formulas in statistics and machine learning include: a) they do not require an explicit form of the probability distributions (or  their density ratio); related probabilistic  quantities can be approximated by statistical estimators over available data; b) they can exploit the  capacity of rich regression models such as neural networks to efficiently search  the  function space for optimal solutions; the optimal solution is typically  related to the  density ratio.

A single divergence  can be derived from several different objective functionals. The key contribution of this paper is a systematic methodology that uses families of transformations  (e.g., shifts, affine, and  powers)  to build new, tighter variational representations for divergences by creating  improved objective functionals, as described in our main  Theorem~\ref{thm:improved_variational_formula_general}.  This idea is both simple and powerful; it provides a general framework that unifies many of the previous variational formulas in the literature, reveals new connections between them,  
drives the derivation of new variational formulas, 
and has practical implications in terms of  accelerated statistical training, learning, and estimation from data.
    
Striking consequences of the proposed framework include: (i) the connection between LT-based KL, the DV representation formula,  and a new, improved  DV-type formula, (ii) a concrete representation of the abstract objective functional in \cite{Ruderman},
and (iii) a derivation of new  representation formulas for $\alpha$-divergence and connections with a recently derived,  
DV-type variational representation of R{\'e}nyi divergences. { Moreover, when the optimization over the transformation family is not analytically tractable, a second-order approximation is employed resulting in new variational representations. 
}

The improved objective functionals  constructed via our framework  have the same optimal solution, but they are tighter in the sense that the same approximation of the optimum will provide a better approximation of the divergence, i.e., they are flatter around the optimal solution.
 %
 %
We   employ 
(functional)  Hessians of the objective functionals   to quantify and order  relative tightness gains between different variational representations of divergences, 
in terms of the local curvature around the optimal solution. { Although our goal here is not variance reduction, we do also study the asymptotic variance of the tighter variational representations. We obtain theoretical   evidence that our optimization procedures do not increase the variance and provide some numerical evidence that the variance as well as the averaged error can be decreased.}


    

    
    
Finally, we  demonstrate that these tighter representation formulas can  accelerate numerical  optimization and estimation of divergences 
in a series of  synthetic and real examples, such as
the statistical  estimation of $f$-divergences and mutual information, including cases with real and/or high-dimensional data (in excess of 1000 dimensions). Similarly to \cite{MINE_paper}, we parameterize the function space  
using neural networks, hence the (parameter) optimization is efficiently performed with back-propagation algorithms. 
{ As an example of  our method, we  develop  neural-based estimators of controlled sample complexity for   Hellinger-based mutual information (Hellinger-MINE). Overall,} we find that  the improved, tighter representation formulas converge several times faster than the initial representation formula, often by nearly an order of magnitude in high-dimensional problems.  

\section{Tightening the Variational Representation of $f$-divergence}
\label{sec:main}
\subsection*{Background}
Define $\mathcal{F}_1(a,b)$ to be the set of all convex functions $f:(a,b)\to \mathbb{R}$ with $f(1)=0$. If $a$ (resp. b) is finite, we extend $f$ to $a$ (resp. $b$) by continuity and set $f(x)=\infty$ for $x\not\in[a,b]$.  Such  functions are appropriate  for defining $f$-divergences, $D_f$, which have the variational characterization 
\begin{align}
D_f(Q\|P)\equiv& E_P[f(dQ/dP)]\notag\\
=&\sup_{\phi\in\mathcal{M}_b(\Omega)}\{E_Q[\phi]-E_P[f^*(\phi)]\}\,,\label{eq:Df_variational_bounded}
\end{align}
where $\mathcal{M}_b(\Omega)$ denotes the set of all bounded measurable functions and $f^*$ is the Legendre transform of $f$
\cite{Broniatowski,Nguyen_Full_2010}. Under appropriate assumptions (see Theorem 4.4 in \cite{Broniatowski}), the maximum is achieved at
\begin{align}\label{eq:optimum_f_divergence}
\phi^*=f^\prime(dQ/dP).
\end{align}
 There are cases, such as $\alpha$-divergence with $\alpha\in (0,1)$, where $D_f(Q\|P)$ can be given a meaningful finite value even if $Q\not\ll P$ (see \cite{LieseVajda}). However, the right hand side of \eqref{eq:Df_variational_bounded} is always $+\infty$ if $Q\not\ll P$ and so here we  use the convention that $D_f(Q\|P)\equiv\infty$ when $Q\not\ll P$.
The special case of Kullback-Leibler (KL) divergence (i.e., $D_f$ with $f(x)=x\log(x)$) has  a well-known alternative variational representation, the Donsker-Varadhan (DV) variational formula
\begin{align}\label{eq:DV}
D_{KL}(Q\|P)=\sup_{\phi\in\mathcal{M}_b(\Omega)}\{E_Q[\phi]-\log E_P[e^\phi]\}.
\end{align}
It is known that the objective functional in \eqref{eq:DV} is tighter than that of \eqref{eq:Df_variational_bounded}, in the sense that $E_Q[\phi]-\log E_P[e^\phi]\geq E_Q[\phi]-E_P[f^*(\phi)]$ for all $\phi \in \mathcal{M}_b(\Omega)$,  \cite{Ruderman}.  
In this paper, we present a general procedure for obtaining tighter variational representations of any $f$-divergence, for which the transition from \eqref{eq:Df_variational_bounded} to \eqref{eq:DV} is just one special case.

\subsection*{Theoretical Results}
Our method for deriving tighter variational representations for $f$-divergences is described in the following Theorem; a proof can be found in Section \ref{sec:L1_variational_formula}.  \\
 
\begin{theorem}\label{thm:improved_variational_formula_general}
 Let $f\in\mathcal{F}_1(a,b)$ and suppose $D_f(Q\|P)<\infty$. With $\phi^*$ defined by \req{eq:optimum_f_divergence} (when it exists), let $\Phi$ be a family of functions (the test functions) with 
\begin{align}\label{eq:Phi_def}
\mathcal{M}_b(\Omega)\subset\Phi\subset L^1(Q) \,\,\text{ or with }\,\,\phi^*\in\Phi\subset L^1(Q).
\end{align}
Consider any  family of transformations 
\begin{align}
    \mathcal{T}\subset\{T=T(\phi)\, , \mbox{ such that }\,  T:\Phi \mapsto  L^1(Q)\} 
\end{align}
that includes the identity map. Then
\begin{align}
D_f(Q\|P)=&\sup_{\phi\in \Phi}H_{\mathcal{T}}[\phi]\label{eq:Df_unbounded_3}\,, \\
 \text{where } \,\,\,\,  H_{\mathcal{T},f}[\phi]=&\sup_{T\in \mathcal{T}}\{
E_Q[T(\phi)]-E_P[f^*(T(\phi))]\}\,,\label{eq:H_def}
\end{align}
and the maximum in \eqref{eq:Df_unbounded_3} is achieved at $\phi^*$. Furthermore, the objective functional, $H_{\mathcal{T},f}$, in the variational representation \eqref{eq:Df_unbounded_3} is   tighter  than  the objective functional in
\eqref{eq:Df_variational_bounded}, in the sense that
\begin{align}\label{eq:T_improvement}
E_Q[\phi]-E_P[f^*(\phi)]\leq 
H_{\mathcal{T},f}[\phi]
\leq
D_f(Q\|P)
\end{align}
for all $\phi \in \Phi$.
\end{theorem}
\begin{remark}
When the choice of $f$ and/or $\mathcal{T}$ is unimportant or clear from the context, we will  omit the corresponding subscript on $H_{\mathcal{T},f}$.
\end{remark}
\noindent
(i) The main new insight and primary mathematical tool in  this paper are formulas \eqref{eq:Df_unbounded_3} and \eqref{eq:H_def}, which allow for the objective functional $H_{\mathcal{T}}[\phi]$ to be  `improved/tightened' using any appropriate  family of transformations $\mathcal{T}$; examples of such families are discussed in the next two subsections. \req{eq:Df_unbounded_3} is a simple but far-reaching idea that reveals  connections between many  known variational representations and also leads to the derivation of new ones. 
(ii) The extension of \req{eq:Df_variational_bounded} from $\mathcal{M}_b(\Omega)$ to $L^1(\Omega)$ is useful because the exact optimizer, $\phi^*$, is generally unbounded; various versions of this extension can be found in the literature \cite{Broniatowski,Nguyen_Full_2010}. This extension is  needed to justify the computation of  variational derivatives around the optimum presented in Section~\ref{sec:Hessians}.
It also implies that one does not need to impose boundedness condition via a cutoff function when employing neural-based statistical estimation. 
(iii) The generalization of \req{eq:Df_variational_bounded} to a family, $\Phi$, that contains the optimizer is the natural next step; again, see \cite{Broniatowski,Nguyen_Full_2010}. It  provides a great deal of flexibility in adapting the proposed variational representation \eqref{eq:Df_unbounded_3} to different $f$-divergences and could guide the algorithmic implementation. We use this idea several times to restrict the optimization to, e.g., positive functions for the $\alpha$-divergences and finite dimensional submanifolds for exponential families of distributions.
%
(iv) If $\mathcal{T}\subset \{T:\Phi\to \Phi\}$  is a group under composition then the objective functional \eqref{eq:H_def} is invariant under the family of transformations $\mathcal{T}$, i.e., $H_{\mathcal{T}}[T(\phi)]=H_{\mathcal{T}}[\phi]$ for all $T\in\mathcal{T}$.  %
(v) If $(\Omega,\mathcal{M})$ is a metric space with the Borel $\sigma$-algebra then one can replace $\mathcal{M}_b(\Omega)$ with $C_b(\Omega)$ (bounded  continuous functions) and $L^1(Q)$ with $L^1_c(Q)$ (continuous  $L^1(Q)$ functions) in Theorem~\ref{thm:improved_variational_formula_general}. This is a direct consequence of Lusin's theorem \cite[Appendix D]{dudley1999uniform}.
(vi) The auxiliary optimization problem in \req{eq:H_def} is often computed analytically; alternatively and  due to its low dimensionality, the corresponding optimization can be easily incorporated into the gradient descent framework without significant additional computational cost.

\subsection*{Families of Transformations}\label{sec:transformations} 
Next, we present  several useful families of transformations that, in conjunction with Theorem \ref{thm:improved_variational_formula_general}, yield tighter variational formulas.

\setlist{nolistsep,leftmargin=*}
\begin{enumerate}[noitemsep]

\item[a.]  Identity: $\mathcal{T}^{{id}}=\{id\}$ leads to what we call the Legendre Transform (LT) $f$-divergence variational formula given by \eqref{eq:Df_variational_bounded}.

\item[b.]  Shifts: $T^{{shift}}_\nu(\phi)=\phi-\nu$, $\nu\in\mathbb{R}$,  lead to what we call the shift or $\nu$-improved variational formula
\begin{align}\label{eq:nu_improved}
D_f(Q\|P)=\sup_{\phi\in \Phi}\{\sup_{\nu\in\mathbb{R}}\{E_Q[\phi]-\nu-E_P[f^*(\phi-\nu)]\}\} \, .
\end{align}
This result was first  obtained in \cite{BenTal2007}, and in this sense Theorem~\ref{thm:improved_variational_formula_general} is a broad and systematic generalization of Theorem 4.4 in \cite{BenTal2007}, where   only shift transformations where considered.

\item[c.] Scaling Transformation: $T^{{scale}}_{\eta}(\phi)=\eta \phi$, $\eta\in\mathbb{R}$, which lead to the scaling or $\eta$-improved variational formula
\begin{align}\label{eq:eta_improved}
D_f(Q\|P)=\sup_{\phi\in \Phi}\{\sup_{\eta\in\mathbb{R}}\{\eta E_Q[\phi]-E_P[f^*(\eta\phi)]\}\} \, .
\end{align}

\item[d.] Affine Transformation: The above two cases can be combined into a two parameter family $T^{affine}_{\eta,\nu}(\phi)=\eta\phi-\nu$. 

\item[e.] Power Transformation: $T_\beta^{power}(\phi)=\phi^\beta$,  $\beta\in\mathbb{R}$, which lead to the power or $\beta$-improved variational formula
\begin{align}\label{eq:beta_improved}
D_f(Q\|P)=\sup_{\phi\in \Phi}\{\sup_{\beta\in\mathbb{R}}\{ E_Q[\phi^\beta]-E_P[f^*(\phi^\beta)]\}\} \, .
\end{align}
As with the affine transformations, the power transformations can be combined with the shift and/or scaling transformations to form a multiparameter family. These are related to the well-known Box-Cox transformation \cite{BoxCox}, used in statistics to transform non-normal data sets to approximately normal.
\end{enumerate}


\subsection*{New Derivations of Existing Variational Representations}

\label{sec:existing:var:rep}

In this and next subsections, we  explore  several specific cases of the above framework, focusing primarily on examples where the optimization over the shift and/or scaling parameter can be done analytically. Here, we uncover connections with previously known variational formulas.

\setlist{nolistsep,leftmargin=*}
\begin{enumerate}[noitemsep]
\item Donsker-Varadhan formula (KL-divergence with shift transformations): KL-divergence is the $f$-divergence corresponding to  $f(x)=x\log(x)$, which has Legendre transform   $f^*(y)=e^{y-1}$. The maximum over the shift transformations $\mathcal{T}^{{shift}}$ in \eqref{eq:Df_unbounded_3} occurs at $\nu^*=\log\left(E_P[e^{\phi}]\right)-1$, and hence 
\begin{align}\label{eq:DV_nu_sup}
   &\sup_{\nu\in\mathbb{R}} \{E_Q[\phi]-\nu-E_P[f^*(\phi-\nu)]\}   \\
   =&E_Q[\phi]-\nu^*-E_P[e^{\phi}]e^{-\nu^*-1}=E_Q[\phi]-\log(E_P[e^\phi]).\notag
\end{align} 
The result is the objective functional in the well-known Donsker-Varadhan variational formula \eqref{eq:DV}
and so this framework provides the connection between \eqref{eq:DV} and  \eqref{eq:Df_variational_bounded}.  We also note that in  \cite{Bridging_fGan_WGan} a  connection is derived between \eqref{eq:DV} and  \eqref{eq:Df_variational_bounded} by using  a logarithmic change of variables in the function space, based on \eqref{eq:optimum_f_divergence} for the KL case.

\item Connection with the results of  \cite{Ruderman}:   In Theorem 1 of \cite{Ruderman} the following improved variational formula was derived: 
 \begin{align}\label{eq:imp_ruderman}
     D_f(Q\|P)=&\sup_{\phi\in\mathcal{M}_b(\Omega)}\{E_Q[\phi]-(I_{f,P}^R)^*[\phi]\},\\
     I_{f,P}^R(r)\equiv&\begin{cases} 
     E_P[f(r)], & r\in L^1(P),\,r\geq 0,\, E_P[r]=1\\
      \infty, &\text{otherwise}
   \end{cases}\,.\notag
 \end{align}
This is another special case of our framework, as can be seen by first rewriting the Legendre-Fenchel transform and then using Theorem 4.2 in \cite{BenTal2007}:
 \begin{align}\label{eq:ruderman_equiv}
     (I_{f,P}^R)^*[\phi]=& \sup_{Q\ll P}\{ E_Q[\phi]- D_f(Q\|P)\}\\
     =&\inf_{\nu\in\mathbb{R}}\{\nu+E_P[f^*(\phi-\nu)]\}.\notag
 \end{align}
Hence the variational formula \eqref{eq:imp_ruderman} is in fact the same as the $\nu$-improved variational formula \eqref{eq:nu_improved}.

\item $\chi^2$-Divergence: 
We use Theorem~\ref{thm:improved_variational_formula_general} to provide a variational perspective on the classical Hammersley-Chapman-Robbins bound for the  $\chi^2$-divergence. To our knowledge, the tightness result of the bound is novel. The $\chi^2$-divergence is a special case of the $\alpha$-divergences, $\chi^2(Q\|P)=2D_{f_2}(Q\|P)$. Here we optimize \eqref{eq:Df_unbounded_3} over $\mathcal{T}^{affine}$ to obtain
\begin{align}\label{eq:HCR}
\chi^2(Q\|P)=& \sup_{\phi\in \Phi}\frac{(E_Q[\phi]-E_P[\phi])^2}{\Var_P[\phi]},
\end{align} 
where $\Phi=\{\phi\in L^1(Q)\cap L^2(P):\phi\geq 0, \Var_P[\phi]>0\}$ and the  maximum is achieved at $\phi^*=dQ/dP\ge 0$.  \req{eq:HCR}  implies  the Hammersley-Chapman-Robbins bound for the $\chi^2$ divergence (see, e.g., Eq. 4.13 in \cite{keener2010theoretical}) and shows tightness over the set $\Phi$. The objective functional in \eqref{eq:HCR} was proposed as loss function for $\chi^2$-GANs, \cite{chi2_Gan}; thus,   \eqref{eq:HCR} provides a complete and  rigorous justification for this choice.
Finally, if we instead optimize  \eqref{eq:Df_unbounded_3} over $\mathcal{T}^{{shift}}$,
we obtain the  objective functional for $\chi^2$ derived in \cite{Bridging_fGan_WGan}, which is thus less tight than \eqref{eq:HCR}; see Appendix \ref{app:alpha_div}.3.

\end{enumerate}

\subsection*{Deriving New Variational Representations and Further Connections}
In this subsection, we produce new variational representations and reveal further connections between divergences. We do not claim these examples are exhaustive. Nevertheless, they cover many important cases and illustrate the power and flexibility of Theorem \ref{thm:improved_variational_formula_general}.

\setlist{nolistsep,leftmargin=*}
\begin{enumerate}[noitemsep]
\item Improved Donsker-Varadhan (KL-Divergence with affine transformations): Introducing a scaling parameter into \req{eq:DV_nu_sup}, i.e., optimizing over all affine transformations
$\mathcal{T}^{affine}$
in \eqref{eq:Df_unbounded_3}, one finds the new KL variational representation
\begin{align}
D_{KL}(Q\|P)
=&\sup_{\phi\in \Phi}\{\sup_{\eta\in\mathbb{R}}\{\eta E_Q[\phi]-\log E_P[e^{\eta\phi}]\}\}\,.
    \, \label{eq:improvedDV}
\end{align}
The inclusions $\mathcal{T}^{{shift}}\subset\mathcal{T}^{affine}$ implies that \eqref{eq:improvedDV} is tighter than DV \eqref{eq:DV}.  Calculations and numerical results that quantify this improved tightness are found in Section \ref{sec:Hessians}. The optimization over $\eta$ in \req{eq:improvedDV} cannot be evaluated analytically in general, but it can be done numerically (as discussed in Section \ref{sec:stat_learning}).

\item $\alpha$-Divergences (scaling transformations):     The $\alpha$-divergences are the family of $f$-divergences corresponding to $     f_\alpha(t)=\frac{t^\alpha-1}{\alpha(\alpha-1)},\,\,\,\alpha\neq 0,1$. See \cite{LieseVajda} 
for  properties, related families, and further references. It includes the  KL, Hellinger and $\chi^2$ divergences as special cases \cite{Minka}, 
is closely related to the Tsallis entropies \cite{Tsallis1988}, 
and appears also in the context of information geometry \cite{amari2016information}.
By optimizing \eqref{eq:Df_unbounded_3} over the family of scaling transformations, $\mathcal{T}^{{scale}}$ (restricted to $\eta>0$), we obtain a  new variational representation of the $\alpha$-divergences:
\begin{align}\label{eq:alpha_div_eta_opt}
    &D_{f_\alpha}(Q\|P) \\
=&\sup_{\phi\in  \Phi_\alpha}\left\{\frac{1}{\alpha(\alpha-1)}\left(E_Q[\phi]^\alpha E_P[\phi^{\alpha/(\alpha-1)}]^{-(\alpha-1)}-1\right)\right\}\,,\notag
\end{align}
where $\Phi_\alpha=\{\phi\in L^1(Q):\phi\geq 0, 0<E_P[\phi^{\alpha/(\alpha-1)}]<\infty\}$ if $\alpha>1$ and $\Phi_\alpha=\{\phi:\phi>0\}$ if $0<\alpha<1$. \req{eq:alpha_div_eta_opt} has the exact optimizers $\phi_\alpha^*=(dQ/dP)^{\alpha-1}$. Theorem \ref{thm:improved_variational_formula_general} guarantees that the objective functionals in the new variational representations \eqref{eq:alpha_div_eta_opt} are tighter than that of the LT $f$-divergence objective functional from \eqref{eq:Df_variational_bounded}. See Appendix \ref{app:alpha_div} for details on the calculations that lead to \req{eq:alpha_div_eta_opt}, as well as for connections  to the KL divergence in the limits as $\alpha\to 0,1$.

\item  Variational representations of R{\'e}nyi divergences: \req{eq:alpha_div_eta_opt} leads to a variational characterization of R{\'e}nyi divergences. Using the known connection between the $\alpha$ and R{\'e}nyi divergences, along with \req{eq:alpha_div_eta_opt}  and the change of variables $\phi=e^{(\alpha-1)g}$ (see Appendix \ref{app:alpha_div} for details) one obtains
\begin{align}\label{eq:Renyi_optim}
    &R_\alpha(Q\|P)=\frac{1}{\alpha(\alpha-1)}\log(\alpha(\alpha-1)D_{f_\alpha}(Q\|P)+1)\\
    =&\sup_{g}\left\{\frac{1}{\alpha-1} \log( E_Q[e^{(\alpha-1)g}]) -\frac{1}{\alpha} \log( E_P[e^{\alpha g}])\right\}\, .\notag
\end{align}
This constitutes an independent derivation of the  Renyi variational formula derived in \cite{doi:10.1137/20M1368926, Bridging_fGan_WGan}, while 
in the asymptotic limit  $\alpha \to  1$ one recovers \eqref{eq:DV}. 
The Renyi variational formula \eqref{eq:Renyi_optim}
was also used in \cite{CumulantGAN:Pantazisetal} to construct cumulant-based, generative adversarial networks.
Moreover, UQ bounds for risk-sensitive functionals in terms of R{\'e}nyi divergences, which were obtained recently  in  \cite{AtarChowdharyDupuis}, readily follow from \eqref{eq:Renyi_optim} after appropriate manipulations and an optimization over $\alpha >1$.

\item $\alpha$-Divergences (scaling and power transformations): For $\alpha\in(0,1)$, Combining the scaling and power family of transformations yields 
\begin{align}\label{eq:Dfalpha_beta_imp}
    &D_{f_\alpha}(Q\|P)=\frac{1}{\alpha(1-\alpha)}\sup_{\phi>0}\sup_{\beta\in\mathbb{R}}\left\{1-E_Q[\phi^\beta]^\alpha E_P[\phi^{-\alpha\beta/(1-\alpha)}]^{1-\alpha}\right\}\,.
\end{align}
The optimization over scalings was evaluated as in \req{eq:alpha_div_eta_opt}
but the optimization over the power transformations, $\beta$, cannot be done analytically. In practice, $\beta$ can be included as an additional  parameter in a numerical optimization procedure; see Section \ref{sec:stat_learning} for further discussion.   

\item Exponential Families: If $P$ and $Q$ are members of a  parametric family then the set of test functions, $\Phi$, in \eqref{eq:Df_unbounded_3} can be reduced to a finite dimensional manifold. For instance, if $P=P_{\theta_p}$ and $Q=P_{\theta_q}$ are members of the same exponential family  $dP_\theta=h(x)e^{\kappa(\theta)\cdot T(x)-\beta(\theta)},\,\,\,\theta\in\Theta$  with $T:\Omega\to\mathbb{R}^n$  the vector of sufficient statistics then the explicit optimizer $\phi^*=f^\prime(dQ/dP)$ lies on an  $(n+1)$-dimensional manifold of functions, parameterized by the sufficient statistics and constants, $\phi_{(\beta,\kappa)}=f^\prime(\exp(\kappa\cdot T+\beta))$, $(\beta,\kappa)\in\mathbb{R}\times\mathbb{R}^n$, and computation of the $f$-divergence reduces to the following finite-dimensional optimization problem:
\begin{align}\label{eq:f_div_manifold}
&D_f(Q\|P)=\sup_{(\beta,\kappa)\in\mathbb{R}^{n+1} }\{E_Q[\phi_{(\beta,\kappa)}]-E_P[f^*(\phi_{(\beta,\kappa)})]\}\,.
\end{align}
This variational representation can be further combined with any appropriate family of transformations
$\mathcal{T}$. We refer to Appendix \ref{app:submanifold} for further details.

\item 
{ Approximating the Improved Donsker-Varadhan: An analytic computation of the optimization over $\eta$ in \req{eq:Df_unbounded_3} is not possible in general. Nevertheless, an alternative variational characterization of the KL divergence can be derived by  expanding around $\eta=1$ and solving the quadratic approximation for $\Delta\eta^*$. Under appropriate assumptions given in Theorem \ref{approx:improved:kld:thm} (see Appendix \ref{app:Imp_DV}), we derive a new variational formula for the KL divergence,
\begin{align}\label{eq:approx_improved_DV}
    &D_{KL}(Q\|P)=\sup_{\phi\in \Phi}\{(1+\Delta\eta^*(\phi))E_Q[\phi]-\log E_P[e^{(1+\Delta\eta^*(\phi))\phi}]\}\, ,
\end{align}
where the optimal $\Delta \eta^*$ is obtained by maximizing the second-order Taylor approximation and it is given by
\begin{align}\label{eq:Delta_eta_star}
\Delta\eta^*(\phi)=\frac{E_Q[\phi]-E_{P_\phi}[\phi]}{\Var_{P_\phi}[\phi]} ,
\end{align}
with $dP_\phi=e^\phi dP/E_P[e^\phi]$ being the tilted measure.
Since we quadratically approximate the improved Donsker-Varadhan, there is no guarantee that the objective functional in \req{eq:approx_improved_DV} is tighter than that of Donsker-Varadhan's representation. Despite no general guarantee, it is expected to be tighter when $\Delta\eta^*(\phi)$ is sufficiently small. This example demonstrates that even when there is no explicit formula for the transformation's optimization, one is still able to derive a closed-form formula for an approximate version of it and thereby obtain a new, rigorous  variational formula. This same methodological approach can be used to analytically approximate the optimization over other families of transformations. As another demonstration, we refer to Theorem \ref{approx:power:renyi:thm} in Appendix \ref{app:Imp_Renyi} where a new approximate variational formula is derived for the power-improved R{\'e}nyi variational representation.}

\item Connections with  Uncertainty Quantification: The improved DV representation \eqref{eq:improvedDV} provides an alternative and arguably more general derivation of model uncertainty bounds derived recently in  \cite{chowdhary_dupuis_2013,DKPP}. These results quantify the effects of model uncertainty by bounding  expectations of an observable $\phi$ under an alternative model, $Q$, in terms of the behavior under a baseline model $P$  and  the model discrepancy, measured by $D_{KL}(Q\|P)$. Specifically, \eqref{eq:improvedDV} implies after straightforward manipulation that
  $ E_Q[\phi] \le \inf_{\eta > 0} \{ \frac{1}{\eta}\log E_P[e^{\eta\phi}]+ \frac{1}{\eta}D_{KL}(Q\|P)\}$  \cite{chowdhary_dupuis_2013}.
More generally, we can obtain similar UQ bounds when model discrepancy is measured by  an $f$-divergence by using 
$T^{affine}_{\eta,\nu}(\phi)=\eta\phi-\nu$ in  \eqref{eq:Df_unbounded_3} and performing the analogous manipulations:
\begin{align}\label{eq:UQ_Df}
E_Q[\phi] \le \inf_{ \substack{\eta>0 \\ \nu \in \mathbb{R}}}\left\{\frac{1}{\eta}\{E_P[f^*(\eta\phi-\nu)]+\nu\}+\frac{1}{\eta}D_f(Q\|P)\right\}\, .
\end{align}
The Hammersley-Chapman-Robbins bound can also be viewed as a special case of \eqref{eq:UQ_Df} in this UQ context.


\end{enumerate}

\section{Variational Derivatives and Tightness
Gains }
\label{sec:Hessians}

\begin{figure}[ht]

\centering
  \includegraphics[width=.48\textwidth]{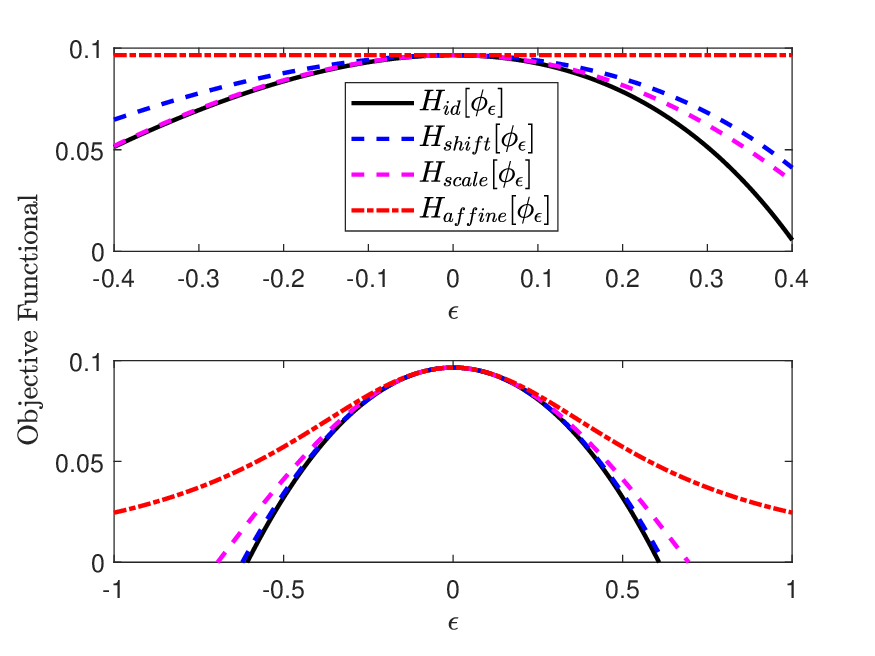}
   \caption{Both plots demonstrate the improvement of the KL divergence objective functional in a neighborhood of the optimizer. Here, $\phi_\epsilon=\phi^*+\epsilon\psi$ where $\psi=x^2$ (top panel) and $\psi(x)=x$ (bottom panel).   $P$ and $Q$ are $1$-dimensional Gaussians. Black curves: LT-based $f$-divergence objective functional, Blue curves: shift-improved (i.e., Donkser-Varadhan), Magenta curves: $\eta$-improved, Red curves: shift-scaling-improved. Note that $\psi=x^2$ is related to $\phi^*$ by a shift and scaling, hence  the shift-scaling-improved objective functional in the top plot has zero curvature in this direction, a manifestation of its shift and scale invariance.}\label{fig:Hessian_test}
 
\end{figure}

In Theorem \ref{thm:improved_variational_formula_general}, we established the general methodology for  building tighter variational representations of  $f$-divergences, by constructing suitable  objective functionals $H_{\mathcal{T}}$. Here we will quantify relative tightness gains corresponding to different transformation families $\mathcal{T}$: for all such families the  maximizer in 
\eqref{eq:Df_unbounded_3} is always  
$\phi^*$ given by \eqref{eq:optimum_f_divergence}. Therefore,  our approach relies on building quadratic variational approximations  of each objective functional \eqref{eq:H_def}
around the common maximizer $\phi^*$,  and subsequently comparing the corresponding (variational) curvatures; see Figure \ref{fig:Hessian_test} for a demonstration.  
Specifically, using that the maximum occurs at $\phi^*$, an asymptotic expansion yields
\begin{align}\label{eq:H_Taylor}
    &H_{\mathcal{T}}[\phi^*+\psi]=D_f(Q\|P)+\frac{1}{2} \langle \nabla^2H_{\mathcal{T}}[\phi^*]\psi,\psi\rangle+O(\|\psi\|^3)\,,
    \end{align}
    where we formally define
   $ \langle \nabla^2H_{\mathcal{T}}[\phi]\psi,\psi\rangle\equiv\frac{d^2}{d\epsilon^2}\big|_{\epsilon=0} H_{\mathcal{T}}[\phi+\epsilon \psi]$
and $\psi$ is any functional perturbation of the maximizer $\phi^*$.
%
%
Formally, the second order term $\nabla^2H_{\mathcal{T}}[\phi^*]$, i.e., a variational Hessian, is necessarily non-positive and determines the behavior in a neighborhood of the maximizer. By comparing 
$\nabla^2H_{\mathcal{T}}[\phi^*]$ for different families $\mathcal{T}$,  we will  quantify the `tightness gains' provided by different transformation families.
All these  calculations can  be made rigorous under appropriate assumptions as demonstrated next in Theorem~\ref{thm:Hessian_KL} and in  Theorem~\ref{thm:Hessian_f_div} in Appendix~\ref{app:Hessian}.

Here we focus our analysis on  affine transformations, $T^{affine}_{\eta,\nu}(\phi)=\eta\phi-\nu$, but a similar analysis can be performed for any family $\mathcal{T}$  with a smooth, finite-dimensional parameterization. 
The LT $f$-divergence variational formula \eqref{eq:Df_variational_bounded} corresponds to $\mathcal{T}$ containing only the identity, and we write the corresponding objective functional as  $H_{{id}}[\phi]$. 
Specializing \eqref{eq:H_def} to the affine case, we  define the  functional
\begin{align}
    H[\phi,\eta,\nu]=&E_Q[T^{affine}_{\eta,\nu}(\phi)]-E_P[f^*(T_{\eta,\nu}^{affine}(\phi))]\\
    =&E_Q[\eta\phi-\nu]-E_P[f^*(\eta\phi-\nu)], \notag
\end{align}
which leads to four different objective functionals and variational representations of the $f$-divergence
 \begin{align}\label{eq:Df_H}
D_f(Q\|P)=&\sup_\phi \overbrace{H[\phi,1,0]}^{H_{{id}}[\phi]}=\sup_{\phi}\overbrace{\sup_{\nu}H[\phi,1,\nu]}^{H_{{shift}}[\phi]}\\
=&\sup_{\phi}\underbrace{\sup_{\eta}H[\phi,\eta,0]}_{H_{{scale}}[\phi]} =\sup_{\phi}\underbrace{\sup_{\eta,\nu}H[\phi,\eta,\nu] }_{H_{affine}[\phi]}\notag
\end{align} 
and the corresponding Hessians $\nabla^2H_{{id}}[\phi]$, $\nabla^2H_{{shift}}[\phi]$, $\nabla^2H_{{scale}}[\phi]$, and $\nabla^2H_{affine}[\phi]$.
%
Next we state a Theorem where we evaluate and compare  these  variational Hessians for the important case of KL divergence, where $f(x)=x\log(x)$ and $\phi^*=f^\prime(dQ/dP)=\log(dQ/dP)+1$. Detailed computations and rigorous analysis for general $f$-divergences  can be found in Appendix \ref{app:Hessian}.

{

Motivated by the formal calculation \eqref{eq:H_Taylor}, we define
\begin{equation}\label{eq:H_functional_epsilon}
    \mathcal{J}_{\mathcal{T}}(\epsilon)=H_{\mathcal{T}}[\phi^*+\epsilon \psi]\, , \quad \mbox{for  any}\quad \psi \in \mathcal{M}_b(\Omega)\quad \mbox{and $\epsilon \in \mathbb{R}$}\, .
\end{equation}
The second derivative, when it exists,  
$\frac{d^2}{d\epsilon^2}\mathcal{J}_{\mathcal{T}}(0)=\frac{d^2}{d\epsilon^2}\big|_{\epsilon=0} H_{\mathcal{T}}[\phi^*+\epsilon \psi]$
is called a 2nd Gateaux derivative of 
$H_{\mathcal{T}}[\phi]$
in the direction $\psi$ and  
describes the curvature of 
the objective functional $H_{\mathcal{T}}[\phi]$  along any given direction  $\psi \in \mathcal{M}_b(\Omega)$ at the maximizer $\phi^*$. Therefore we will define a corresponding variational Hessian   in the direction $\psi$ as 
\begin{align}\label{eq:Hessian_def}
   \langle \nabla^2H_{\mathcal{T}}[\phi]\psi,\psi\rangle\equiv
   \frac{d^2}{d\epsilon^2}\mathcal{J}_{\mathcal{T}}(0)=\frac{d^2}{d\epsilon^2}\big|_{\epsilon=0} H_{\mathcal{T}}[\phi+\epsilon \psi]\, .
   \end{align}
We remark that    relation \eqref{eq:Hessian_def} is simply a notation that only intends to draw a parallel to finite dimensional optimization calculations using conventional Hessians at maxima or minima of finite dimensional functions. All rigorous results are stated in terms of the 2nd Gateaux derivative $\frac{d^2}{d\epsilon^2}\mathcal{J}_{\mathcal{T}}(0)$.
%
%
Using the above notations we can formulate  to following result  for  the KL divergence.
\begin{theorem}[Tightness gains for KL divergence]
\label{thm:Hessian_KL} Consider the KL divergence, i.e., $f(x)=x\log(x)$.
In addition to the assumptions of Theorem~\ref{thm:improved_variational_formula_general}, we assume that the maximizer $\phi^*=\log(dQ/dP)+1$ in  \eqref{eq:optimum_f_divergence} is bounded (i.e., $\phi^* \in \mathcal{M}_b(\Omega)$) and  select  
the function space $\Phi= \mathcal{M}_b(\Omega)$ in 
Theorem~\ref{thm:improved_variational_formula_general}.
Then for  $\, \mathcal{T}= \mathcal{T}^{{id}}\, ,
   \mathcal{T}^{{shift}}\, ,
   \mathcal{T}^{affine}$
the function
$\mathcal{J}_{\mathcal{T}}$ in \eqref{eq:H_functional_epsilon} is twice differentiable 
at $\epsilon=0$. Furthermore, using the notation 
\eqref{eq:Hessian_def},
the corresponding 2nd Gateaux derivatives $\frac{d^2}{d\epsilon^2}\mathcal{J}_{\mathcal{T}}(0)=
\frac{d^2}{d\epsilon^2}\big|_{\epsilon=0}
H_{\mathcal{T}}[\phi^*+\epsilon \psi]$ 
for all   $\mathcal{T}= \mathcal{T}^{{id}}\, , \mathcal{T}^{{shift}}\, , \mathcal{T}^{affine}$ satisfy the following:
\begin{align}
   &\frac{d^2}{d\epsilon^2}\mathcal{J}_{\mathcal{T}^{{id}}}(0) =\langle\nabla^2H_{{id}}[\phi^*]\psi,\psi\rangle=-\Var_Q[\psi] - E_Q[\psi]^2\,,
    \label{H_unoptimized}\\
   &\frac{d^2}{d\epsilon^2}\mathcal{J}_{\mathcal{T}^{{shift}}}(0) =\langle \nabla^2 H_{{shift}}[\phi^*]\psi,\psi\rangle=-\Var_Q[\psi]\,,
    \label{eq:hessian:gain:shift}\\
&\frac{d^2}{d\epsilon^2}\mathcal{J}_{\mathcal{T}^{affine}}(0)=\langle \nabla^2H_{affine}[\phi^*]\psi, \psi\rangle\label{eq:hessian:gain:affine}=-\Var_Q[\psi] + \frac{\cov_Q(\phi^*,\psi)^2}{\Var_Q[\phi^*]}\,,
\end{align}
corresponding to \eqref{eq:Df_variational_bounded}, \eqref{eq:DV}, and \eqref{eq:improvedDV}, respectively. 
\end{theorem}

The complete proof of Theorem \ref{thm:Hessian_KL} is presented in Appendix~\ref{app:Hessian} in the form of   Theorem~\ref{thm:Hessian_f_div} that describes the  more general case of $f$-divergences. 
\begin{remark}
The  gains inherent in the inclusions $\mathcal{T}^{{id}}\subset\mathcal{T}^{{shift}}\subset\mathcal{T}^{affine}$ in Theorem \ref{thm:improved_variational_formula_general} are quantified in Theorem~\ref{thm:Hessian_KL} by comparing the variational curvatures \req{H_unoptimized}, \req{eq:hessian:gain:shift}, and \req{eq:hessian:gain:affine} as computed by these 2nd Gateaux derivatives; note that they are progressively smaller in magnitude. These curvature computations demonstrate how one can rigorously and precisely quantify  heuristics such as those presented in Figure 1 from \cite{Ruderman}. Furthermore, our  Hessian computations in the form of   Theorem~\ref{thm:Hessian_f_div} also  quantify and extend  to $f$-divergences the accuracy gains observed in the neural estimation of mutual information in  \cite{MINE_paper}.
\end{remark}
\begin{remark}
The boundedness assumptions of Theorems~\ref{thm:Hessian_KL}
and \ref{thm:Hessian_f_div} may appear restrictive compared to the generality assumed for $f$-divergence definition, but they provide the simplest conditions under which  Theorems can be rigorously stated. Under appropriate technical assumptions these results can be easily generalized without assuming the boundedness of $\phi^*$ and for more general function spaces $\Phi$.
Any necessary  assumptions  need to  ensure the applicability of the implicit function theorem and the dominated convergence theorem \cite{Folland} (Theorem 2.27), following their use in the proof of Theorem~\ref{thm:Hessian_f_div} presented in Appendix~\ref{app:Hessian}.
\end{remark}

}

Figure \ref{fig:Hessian_test} is a simple  demonstration  of  Theorem~\ref{thm:Hessian_KL}, using 1D Gaussians: $P=N(0,1)$, $Q=N(0,1/2)$, with perturbations in the directions $\psi=x^2$ (top) and $\psi=x$ (bottom). Optimizing over all affine transformations (red curves) provides noticeable curvature gains when compared to optimization over only  shifts (blue curves), i.e., the improved DV proposed in  \eqref{eq:improvedDV} compared to the classical DV objective functional \eqref{eq:DV} and even more so compared to the Legendre transform case,  \eqref{eq:Df_variational_bounded} (black curves).

{
\section{$f$-Divergence Estimator Bias and Variance}\label{sec:var}

The variational formula \req{eq:Df_unbounded_3} suggests the following natural $f$-divergence estimator
\begin{align}\label{eq:Df_estimator}
    \widehat{D}^n_f(Q\|P)=\sup_{\phi\in\widetilde{\Phi}}\sup_{T\in\mathcal{T}}\{E_{Q_n}[T(\phi)]-E_{P_n}[f^*(T(\phi))]\}\,,
\end{align}
where $Q_n$ and $P_n$ are the $n$-sample empirical measures and $\widetilde{\Phi}\subset\Phi$ is a function space that can be optimized over numerically (e.g., a family of neural networks). A natural question is therefore the  bias and variance of this estimator. In practice, the optimizations are performed via some stochastic gradient descent (SGD) algorithm, and one is actually interested in the bias and variance after a finite number of training steps. Addressing this complicated problem, which depends heavily on the choice of space $\widetilde{\Phi}$ and the SGD algorithm, is outside the scope of the present work. However, in this section we will follow the prior work in \cite{Song2020Understanding} for KL divergences and discuss the bias and variance of the objective functional in \eqref{eq:Df_estimator}. Finally, we emphasize that our goal of this paper is to develop tighter objective functionals and study the impact of improved curvature on the speed of convergence of numerical estimators.  This is a separate question from that of variance reduction, which we do not pursue here. However, we will show that, for $\alpha$-divergences, optimizing over scalings does not worsen the variance at the optimizer.

\subsection*{ Objective Functional Bias}
The estimator \eqref{eq:Df_estimator} can be viewed either as a single-stage optimization problem over both $\phi$ and $T$, in which case the objective functional is unbiased, or as a two-stage optimization,
\begin{align}
    \widehat{D}^n_f(Q\|P)=&\sup_{\phi\in\widetilde{\Phi}} \widehat{H}_{\mathcal{T},f}[\phi;Q_n,P_n]\,,\\
    \widehat{H}_{\mathcal{T},f}[\phi;Q_n,P_n]=&\sup_{T\in\mathcal{T}}\{E_{Q_n}[T(\phi)]-E_{P_n}[f^*(T(\phi))]\}\,,\notag
\end{align}
with a biased objective functional. If the optimization over $T$ must be performed numerically then we take the former view, but if part of the optimization  can be performed analytically (such as in \eqref{eq:DV_nu_sup} and \eqref{eq:alpha_div_eta_opt}) then we take the latter view. In any case, the  divergence estimator as a whole, \req{eq:Df_estimator}, is not unbiased. This can be seen in the simple case of discrete measures on a finite sample space, where one can numerically optimize over all of $\mathcal{M}_b(\Omega)$,
\begin{align}
    \widehat{D}^n_f(Q\|P)=&\sup_{\phi\in\mathcal{M}_b(\Omega)}\{E_{Q_n}[\phi]-E_{P_n}[f^*(\phi)]\}
    =D_f(Q_n\|P_n)\notag\\
    =&E_{P_n}[f(dQ_n/dP_n)]\,,
\end{align}
which is biased. This fact renders the biased/unbiased objective functional question less relevant in practice, as one's goal is generally to estimate the divergence and not just the objective functional at a fixed $\phi$. 

\subsection*{ Objective Functional Variance}

For general $f$, the objective functional estimator for the   LT variational formula is
\begin{align}
\widehat{H}_f[\phi;Q_n,P_n]=E_{Q_n}[\phi]-E_{P_n}[f^*(\phi)]\,.    
\end{align}
The two terms are independent, therefore
\begin{align}\label{eq:LT_variance}
    \Var[\widehat{H}_f[\phi;Q_n,P_n]]=&\Var[E_{Q_n}[\phi]]+\Var[E_{P_n}[f^*(\phi)]]\\
    =&\frac{1}{n}\Var_Q[\phi]+\frac{1}{n} \Var_P[f^*(\phi)]\,.\notag
\end{align}
In particular, for $\alpha$ divergences \eqref{eq:LT_variance} reduces to
\begin{align}\label{eq:LT_alpha_var}
 &   n\Var[\widehat{H}_{f_\alpha}[\phi;Q_n,P_n]]=\Var_Q[\phi]+\alpha^{-2}|\alpha-1|^{2\alpha/(\alpha-1)} \Var_P[\phi^{\alpha/(\alpha-1)}]\,,
\end{align}
where for $\alpha\in(0,1)$ we made a change of variables $\phi\to-\phi$.

To compute the asymptotic variance of an optimized objective functional, we rely on   the delta method. This method can be applied to any objective functional that can be expressed as a function of expectations.  We provide the details in the case of scaling-optimized $\alpha$-divergences \req{eq:alpha_div_eta_opt}. 
\begin{theorem}\label{thm:asymptotic_var} Let   $\alpha>0$, $\alpha\neq 1$ and suppose $\phi\in\Phi$    satisfies $0<c\leq\phi\leq d$ for some $c,d\in\mathbb{R}$.  Then we have:

 (a)  The scaling-optimized $\alpha$-divergence objective functional \eqref{eq:alpha_div_eta_opt},
\begin{align}\label{eq:H_alpha_scale}
&\widehat{H}_{scale,f_\alpha}[\phi;Q_n,P_n]=\frac{1}{\alpha(\alpha-1)}(E_{Q_n}[\phi]^\alpha E_{P_n}[\phi^{\alpha/(\alpha-1)}]^{-(\alpha-1)}-1)\,,
\end{align}
has asymptotic variance
 \begin{align}\label{eq:asymp_var_thm}
        &\lim_{n\to\infty}n\Var[\widehat{H}_{scale,f_\alpha}[\phi;Q_n,P_n]]\\
        =&(\alpha-1)^{-2} (E_Q[\phi]/E_P[\psi_\alpha])^{2(\alpha-1)}\Var_Q[\phi]+\alpha^{-2}(E_Q[\phi]/E_P[\psi_\alpha])^{2\alpha}\Var_P[\psi_\alpha])\,,\notag
    \end{align}
    where $\psi_\alpha\equiv\phi^{\alpha/(\alpha-1)}$.

    (b) At the exact optimizer, $\phi^*_\alpha=(dQ/dP)^{\alpha-1}$, \req{eq:asymp_var_thm} reduces to 
\begin{align}\label{eq:alpha_var_optimizer}
&\lim_{n\to\infty}n\Var[H_{scale,f_\alpha}[\phi^*;Q_n,P_n]] \\
   =&|\alpha-1|^{-2}\Var_Q[(dQ/dP)^{\alpha-1}]+\alpha^{-2}\Var_P[(dQ/dP)^\alpha]\,.\notag
\end{align}
\end{theorem}
See   Appendix \ref{app:asymp_var} for the proof.

Theorem~\ref{thm:asymptotic_var} implies that we need $(dQ/dP)^{2\alpha}\in L^1(P)$
to achieve a finite variance; this fact can motivate an appropriate choice of $\alpha$; we also refer to the Hellinger-MINE discussion and Corollary~\ref{corr:Hellinger:variance} in  Section~\ref{sec:stat_learning}. The asymptotic variance \eqref{eq:alpha_var_optimizer} agrees with the variance of the LT objective functional, \req{eq:LT_alpha_var}, at its optimizer $\phi_\alpha^*=(dQ/dP)^{\alpha-1}/|\alpha-1|$. Hence, from the perspective of the variance at the optimizer, neither method has an advantage. Away from the optimizer there is no consistent relationship between the two variances. In practice,  both methods will take different paths to the optimizer and this further complicates any variance comparison away from the optimizer.  However, empirically we found that the estimators constructed via the improved variational formulas have smaller variance at the estimated optimal which is different in general from the theoretical optimizer $\phi^*$ (i.e., after a finite number of SGD steps); see Figure \ref{fig:Hellinger:embed:error}. }

\section{Numerical Examples: Faster Statistical Estimation and Learning}
\label{sec:stat_learning} 
Next we discuss   practical  implications of using  tighter variational representations developed in Theorem~\ref{thm:improved_variational_formula_general}, focusing    on  accelerating   neural-based statistical learning and estimation. 
In recent works, variational representations such as \eqref{eq:Df_variational_bounded} or \eqref{eq:DV} were used to estimate $f$-divergences and     likelihood ratios 
based solely on available data \cite{Nguyen_Full_2010}. This variational perspective proved also to be a crucial mathematical step  in training generative adversarial networks (GAN) \cite{GAN,f_GAN,WGAN} and towards developing neural-based estimators for mutual information, \cite{MINE_paper}, taking advantage  of the ability of neural networks  to search efficiently through function spaces.

Improved variational formulas for statistical estimation and learning were  previously studied in:  i) \cite{Ruderman}, using \req{eq:imp_ruderman} and assuming a Hilbert space (RKHS) function space, ii) \cite{MINE_paper}, where  the DV and LT formulas for the KL divergence were used to estimate mutual information with improved accuracy. Both of these implicitly rely on the shift-improved variational formula (see \req{eq:DV_nu_sup} and \req{eq:ruderman_equiv}).  Our  Theorem \ref{thm:improved_variational_formula_general} provides a broad generalization of these ideas to other transformation families, allows for practical implementation of the method in \cite{Ruderman} to more general functions space parametrizations (e.g., neural networks), and generalizes the ideas in \cite{MINE_paper} to other $f$-divergences beyond KL, where it can provide improved mutual information estimators based on \eqref{eq:Df_unbounded_3}.  

In the following, we employ the outcomes from Sections \ref{sec:main} \& \ref{sec:Hessians} and build  several variational neural network estimators, in the general spirit of \cite{Nguyen_Full_2010,MINE_paper}. We demonstrate the  performance improvements that result from  representations such as  \req{eq:Df_unbounded_3}.
We start  with the heuristic observation, illustrated in Figure~\ref{fig:Hessian_test}, that tighter representations can improve the accuracy of statistical estimators for $f$-divergences,  in the sense that the same approximation of the optimal $\phi^*$ will provide a better approximation of the divergence. Moreover, tighter variational formulas can lead to faster convergence of the search algorithm, as we now motivate: suppose one minimizes a convex function $f(x)$ by the simple gradient descent algorithm  $x_{n+1}=x_n-\gamma \nabla f(x_n)$. If $\nabla f$ is $L$-Lipschitz (i.e., the Hessian is bounded by $L$) then this algorithm converges if $0 < \gamma< 2/L$, and  the analysis suggests the optimal learning rate of $\gamma=1/L$ and leads to the error bound $|f(x_n)-f(x^*)|\leq \|x_0-x^*\|^2/(2\gamma n)$ (see, e.g., Theorem 3.3 in \cite{Bubeck}). If $\widetilde f$ has the same optimizer and optimal value, but has a smaller  Hessian bound, $\widetilde L$,  then the optimal learning rate, $\widetilde\gamma=1/\widetilde{L}$ is larger,  and the error bound after an equal number of steps is smaller, i.e., the use of $\widetilde{f}$ in place of $f$ can lead to faster convergence.

The above argument is only heuristic; the constant learning rate algorithm is far from optimal in most cases and the above analysis does not  capture the complexity of the current setting. Nonetheless, it does provide important insight into the numerical results, presented below,  which demonstrate that, in practice, the improved variational formulas do generally lead  to faster convergence of the estimators,  letting all other factors be equal.

The  examples below use the new variational formulas \eqref{eq:alpha_div_eta_opt} and \eqref{eq:Dfalpha_beta_imp} for $\alpha$-divergence, largely focusing on the well-known Hellinger divergence ($\alpha=1/2$), defined as 
\begin{align}
D_{f_{1/2}}(Q\|P)=&4\left(1-\int p^{1/2}q^{1/2}d\mu\right)\,,
\end{align}
where $dQ=qd\mu$, $dP=pd\mu$.  Computations were done in TensorFlow using the AdamOptimizer \cite{kingma2014adam}, an adaptive learning-rate SGD optimizer, with all methods given the same initial learning rate.  When working with neural-network based estimators of $\alpha$-divergence, we enforce positivity of the test functions (see \eqref{eq:alpha_div_eta_opt} and \eqref{eq:Dfalpha_beta_imp}) via the parameterization $\phi_\theta=\exp(g_\theta)$ where $g_\theta$, $\theta\in\Theta$, is a neural network  family with ReLU activation functions. We compare the LT method, \eqref{eq:Df_variational_bounded} with the improved estimators based on our formulas \eqref{eq:alpha_div_eta_opt} and \eqref{eq:Dfalpha_beta_imp}.  For instance, in the case of Hellinger divergences we  compare the following estimators.\\
{\bf LT Hellinger Estimator: }
\begin{align}
  \sup_{\theta\in\Theta}\{E_{Q_N}[-\exp(g_\theta)]- E_{P_N}[f_{1/2}^*(-\exp(g_\theta))]\}\,,
\end{align}
\begin{align}
f_{1/2}^*(y)=\infty 1_{y\geq 0}+4(|y|^{-1}-1)1_{y<0}\,,
\end{align}
{\bf Scaling-Improved Hellinger Estimator: }
\begin{align}
4\sup_{\theta\in\Theta}\left\{1-E_{Q_N}[\exp(g_\theta)]^{1/2} E_{P_N}[\exp(-g_\theta)]^{1/2}\right\}\,,
\end{align}
{\bf Scaling-Power-Improved Hellinger Estimator: }
\begin{align}
4\sup_{\theta\in\Theta,\beta\in\mathbb{R}}\left\{1-E_Q[\exp(\beta g_\theta)]^{1/2} E_P[\exp(-\beta g_\theta)]^{1/2}\right\}\,.
\end{align}
In the above, $Q_N$ and $P_N$ denote the expectation under the empirical distributions using $N$ iid samples from $Q$ and $P$ respectively.

{
If the optimization over a parameterized family of transformations, $T_{\delta}$, cannot be performed analytically then we solve the minimization problem \eqref{eq:Df_unbounded_3} - \eqref{eq:H_def} by performing stochastic gradient descent (SGD) on the full collection of parameters, $(\theta,\delta)$.   In such cases, our two-step formulation can be thought of as parametric enhancement of the neural network architecture.  The nested nature of the minimization over $\phi$ and $T$  also allows for more sophisticated methods (not explored here), e.g., for each $\phi$ one can perform several SGD steps for $T$, thus solving the (generally low dimensional) problem \eqref{eq:H_def} to high accuracy, before performing another  SGD step for $\phi$ in \eqref{eq:Df_unbounded_3}; this is reminiscent of multiscale numerical methods \cite{multiscale}. The parameterization,  $\delta$, of the families of transformations considered here  is at most two dimensional. For example in the case of $T^{affine}$ we have $\delta=(\nu, \eta)$, where $ \nu,\eta \in \mathbb{R}$; see also the remaining examples in Section~\ref{sec:transformations}. Including this small number of additional parameters in the stochastic gradient descent iterations is expected to add a negligible additional computational cost, as compared to the (generally) much larger number of neural-network parameters, $\theta$.  In practice, we do find the additional computational cost to be negligible. 
}

\subsection*{Hellinger-MINE}
Here we consider the problem of computing  Hellinger mutual information (Hellinger-MI), $D_{f_{1/2}}(P_{(X,Y)}\|P_X\times P_Y)$.
Typically the divergence in mutual information $D_{f}(P_{(X,Y)}\|P_X\times P_Y)$ is chosen to be KL. However, one can consider a whole array of different $f$-divergences for this purpose, see for instance \cite{f-sensitivity}.  { A motivation for choosing Hellinger over KL is rigorously based on the  variance calculations in Section \ref{sec:var}. In particular, as a direct consequence of Theorem~\ref{thm:asymptotic_var} we obtain the following:
\begin{corollary}\label{corr:Hellinger:variance}
Under the assumptions of Theorem~\ref{thm:asymptotic_var},  the relative variance for the Hellinger $\alpha$-divergence where $\alpha=1/2$, at the optimizer $\phi^*$ is
\begin{align}\label{eq:Hellinger_asymp_var}
    \frac{n\Var[\widehat H_{f_{1/2}}^n[\phi^*;Q,P]]}{D_{f_{1/2}}(Q\|P)^2}=&\frac{\lim\limits_{n\to\infty}n\Var[\widehat H_{scale,f_{1/2}}[\phi^*;Q_n,P_n]]}{D_{f_{1/2}}(Q\|P)^2}\\
    =&\frac{8- D_{f_{1/2}}(Q\|P)}{2 D_{f_{1/2}}(Q\|P)}\,.\notag
\end{align}
Therefore, the sample complexity of the estimator $\widehat H_{scale,f_{1/2}}[\phi^*;Q_n,P_n]$ at the  optimizer $\phi^*$ is $n=\mathcal{O}(1)$, when $Q\ne P$.
\end{corollary}
Comparing Corollary~\ref{corr:Hellinger:variance} with the corresponding result for the KL divergence from Theorem 2 of Ref. \cite{Song2020Understanding},
 \begin{align}
    \frac{\lim\limits_{n\to\infty}n\Var[ \widehat H_{DV}[\phi^*;Q_n,P_n]]}{D_{KL}(Q\|P)^2}\geq&  \frac{e^{D_{KL}(Q\|P)}-1}{D_{KL}(Q\|P)^2}\,,
    \end{align}
 we see that in practice the KL divergence  requires $n=\mathcal{O}(e^{D_{KL}( Q, P)})$ samples, while the Hellinger divergence requires     $n=\mathcal{O}(1)$ samples due to \eqref{eq:Hellinger_asymp_var}.}

\begin{figure}[ht]
  \centering
  \includegraphics[width=.48\textwidth]{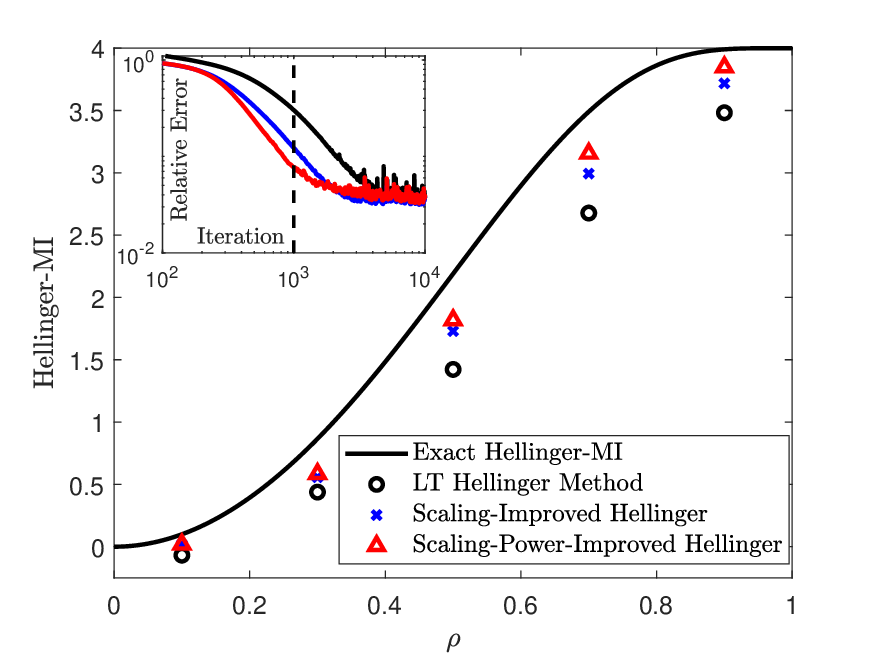}
   \caption{Estimation of Hellinger-based mutual information  between  $20$-dimensional correlated Gaussians with component-wise correlation $\rho$. We use a fully-connected neural network with one hidden layer of 64 nodes while training is performed with a minibatch size of 100. We show the Hellinger MI as a function of $\rho$ after 1000 steps of SGD and averaged over 50 runs. The inset shows the relative error  for $\rho=0.7$, as a function of the number of SGD iterations. }\label{fig:hellinger_MINE}
\end{figure}

In Figure \ref{fig:hellinger_MINE}, we present the computation of Hellinger mutual information (Hellinger-MI), $D_{f_{1/2}}(P_{(X,Y)}\|P_X\times P_Y)$, via neural network optimization,  where $X$ and $Y$ are correlated  $20$-dimensional Gaussians  with component-wise correlation $\rho$.   The results demonstrate that, for a given computational budget (i.e., fixed number of SGD iterations) the improved variational formulas (red and blue) yield more accurate results, i.e., they converge faster than the LT $f$-divergence method \eqref{eq:Df_variational_bounded} (black). Moreover,  optimizing over both scalings and powers \eqref{eq:Dfalpha_beta_imp} (red) provides a non-trivial improvement over the scaling-improved method \eqref{eq:alpha_div_eta_opt}  (blue).  This is a generalization of the findings in \cite{MINE_paper}, which compared the DV variational formula \eqref{eq:DV} with \eqref{eq:Df_variational_bounded}   for the KL divergence. We emphasize that despite the lack of an analytical formula for the optimization over $\beta$, the inclusion of this single additional parameter in the variational formula (a negligible addition to the computational cost) leads  to a clear performance gain.

\begin{figure}[ht]
\begin{minipage}[b]{0.49\linewidth}
\centering
\includegraphics[width=\textwidth]{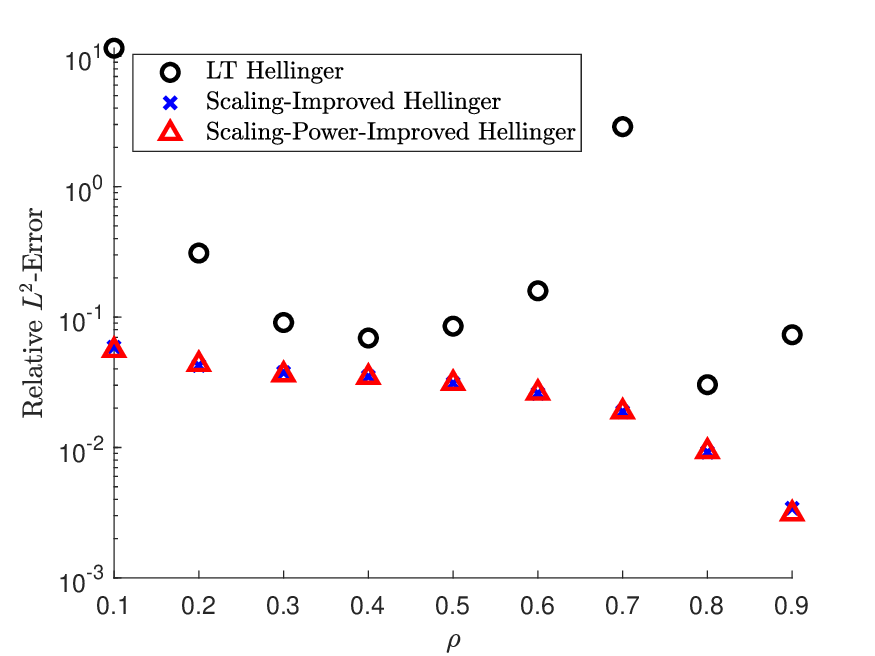}
\end{minipage}
\begin{minipage}[b]{0.49\linewidth}
\centering
\includegraphics[width=\textwidth]{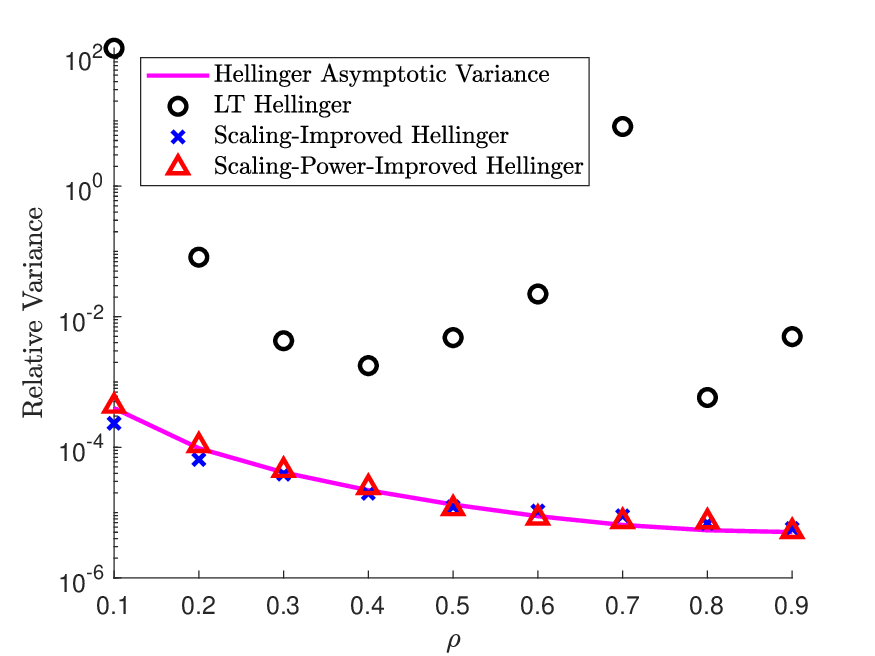}
\end{minipage}

\begin{minipage}[b]{0.49\linewidth}
\centering
\includegraphics[width=\textwidth]{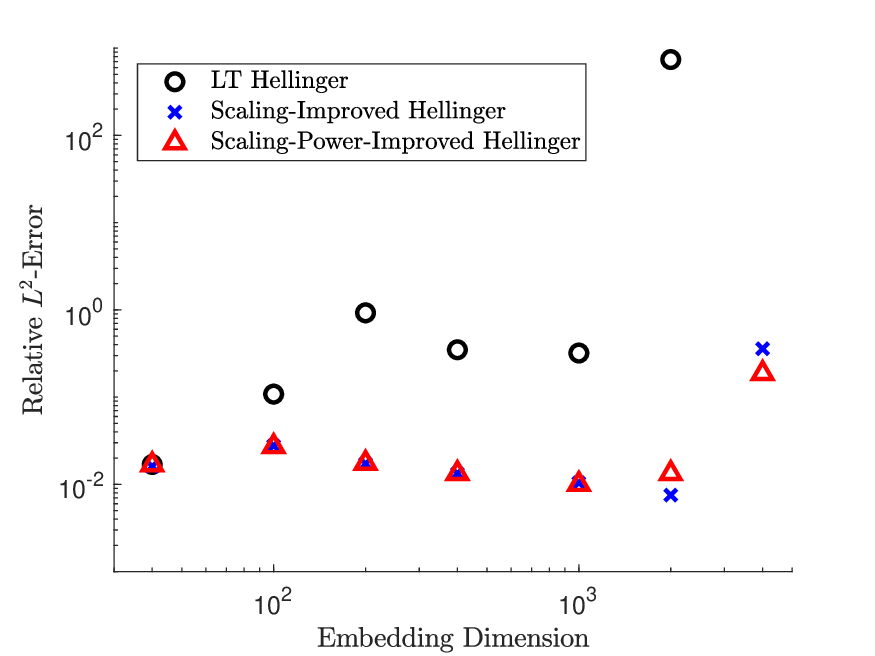}
\end{minipage}
\begin{minipage}[b]{0.49\linewidth}
\centering
\includegraphics[width=\textwidth]{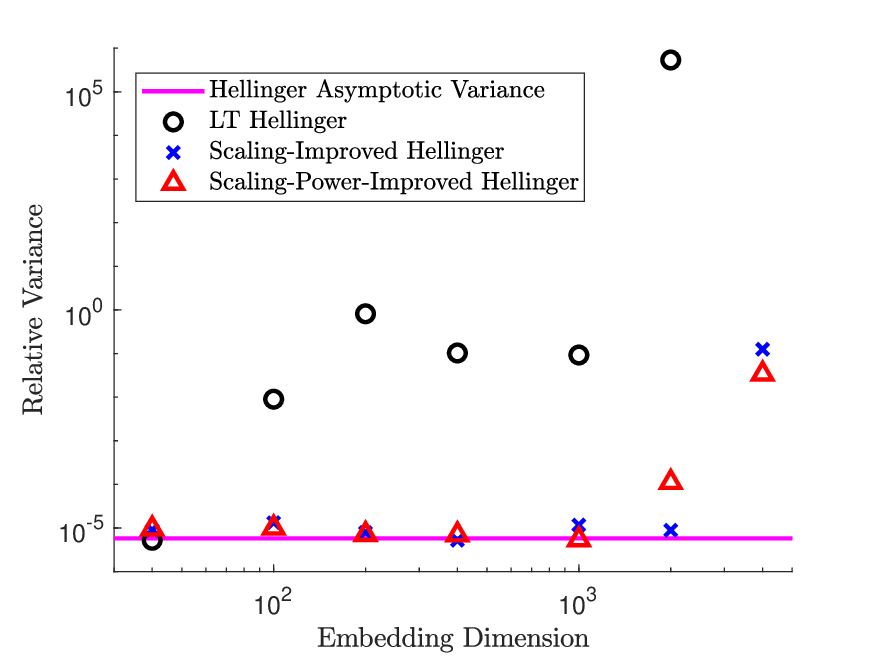}
\end{minipage}
 \caption{{ Hellinger-MINE between correlated Gaussians embedded in high-dimensional space by a map of the following form: $h_i(x)=x_i$ for $i=1,...,20$ and for $i>20$ we  define $h_i(x)=A_i(x)+\cos(x_{j_{1,i}})\sin(x_{j_{2,i}})+x_{j_{3,i}}x_{j_{4,i}}$, 
   where $A$ is an affine function  and $j_{k,i}\in\{1,...,20\}$; the parameters of $A$ were randomly selected. We employ a fully-connected neural network  and performed training with a minibatch size of 1000 from a dataset of 100000 samples. We show results after 50000 steps of SGD and the relative $L^2$-error was computed using data from 75 runs. The left figures show the relative error while the right figures show the relative variance; the solid lines show the exact relative asymptotic variance \eqref{eq:Hellinger_asymp_var}.  The top row show the result as a function of $\rho$ for embedding-dimension $200$ and used a single hidden layer of 64 nodes. The bottom row shows the result for $\rho=0.75$ as a function of problem dimension (i.e., embedding dimension for $X$ plus embedding dimension for $Y$) and used two hidden layers of 64 and 8 nodes respectively, irrespective of the embedding dimension; the majority of LT method runs diverged when the dimension equaled $4000$. } }
\label{fig:Hellinger:embed:error}
\end{figure}

{ As a followup, we demonstrate the effectiveness of our method in estimating Hellinger-MI for high-dimensional problems with low-dimensional structure.  Specifically, in Figure \ref{fig:Hellinger:embed:error} we compare 20-dimensional Gaussians embedded in high dimensional space via a nonlinear map.  The left panels  demonstrate the  performance gain when using  the curvature-improved objective functionals; we find that the optimized methods significantly outperform the LT method, especially in higher dimensions.  The right panels are a (partial) demonstration that the variances are comparable, if not improved, when using the optimized objective functionals. Specifically, we find that the optimized methods approach the exact asymptotic variance faster while the LT method, which  has not yet converged, has a larger variance.}

\subsection*{Submanifold Parameterization for Exponential Families} 

\begin{figure}[ht]
  \centering
\includegraphics[width=.48\textwidth]{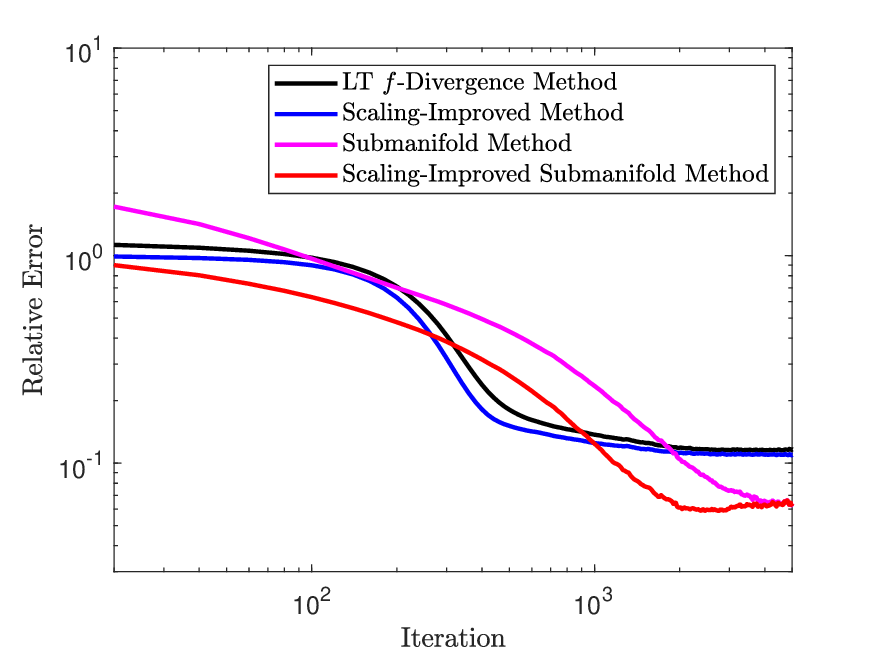}  
\caption{Estimation of the $\alpha$-divergence with $\alpha=0.25$ between two $10$-dimensional Gaussians with randomly generated variances and one of the means randomly perturbed from zero. We compare the convergence performance between the neural network and submanifold parameterizations. The relative error was averaged over 50 runs.}
  \label{fig:hellinger_submanifold}
  \end{figure}

Our method allows for a great deal of flexibility in the choice of function space parameterization. In a `small-data' setting, the assumption of an exponential family structure can serve as an effective regularization.  We illustrate this with Figure \ref{fig:hellinger_submanifold}, which  shows the estimation of the $\alpha$-divergence with $\alpha=0.25$ between $10$-dimensional Gaussians using a data set of  5000 samples from each distribution for SGD (minibatch size of 100) and using another 5000 samples for Monte Carlo estimation of the value of the objective functional.  Using  the submanifold estimation formula  \eqref{eq:f_div_manifold} and its scaling-improved variant (see Appendix \ref{app:submanifold})  we obtain the magenta and red curves, respectively.   In   blue, we show the result from the scaling-improved variational formula \eqref{eq:alpha_div_eta_opt} and in black we show the result using the LT $f$-divergence objective functional \eqref{eq:Df_variational_bounded};
  both use neural network families with one fully connected hidden layer (5 nodes). The number of nodes  was chosen so that all methods use approximately the same number of parameters. The neural network parameterization converges faster, but ends up with  a larger bias than the submanifold parameterization. The scaling-improved variational formulas lead to faster convergence than the LT variational formula in both cases as expected by our theory.

\subsection*{MNIST Dataset Examples} 
\begin{figure}
\centering
  \includegraphics[width=.48\textwidth]{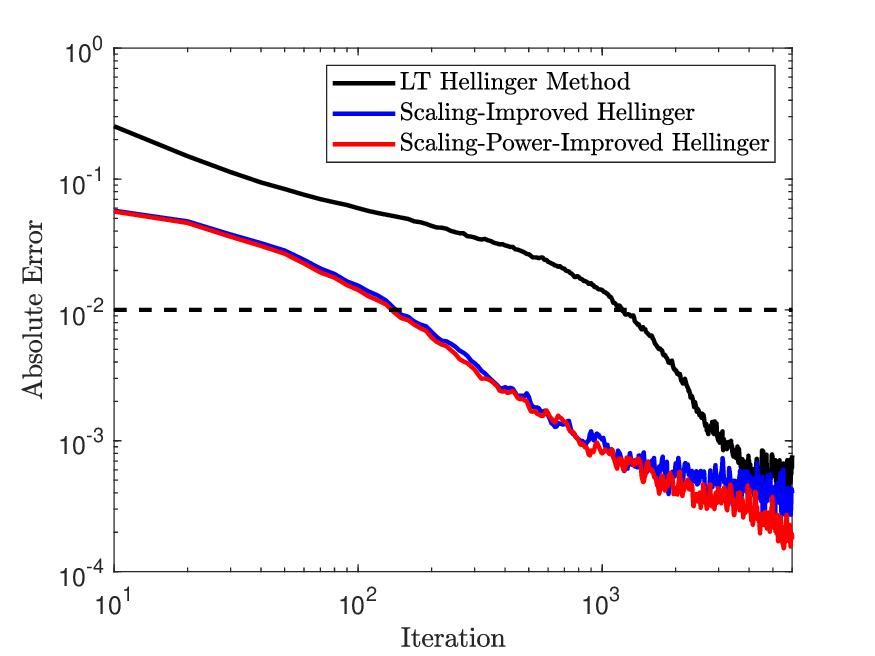}
  \caption{Estimation of the Hellinger divergence  between two distributions obtained by (iid) randomly translating the MNIST handwritten digits dataset \cite{MNIST}: each sample  is a random translation of an MNIST image in the $x$ and $y$ directions (iid $N(0,3^2)$ shifts, rounded to the nearest pixel and with periodic boundary conditions). Each step of SGD uses two independent minibatches of 100 such samples (one minibatch for $P$ and one for $Q$). Monte Carlo estimation of the value of the Hellinger divergence was done  using the corresponding objective functional and with samples coming from  two separate  datasets of 10000 randomly  shifted MNIST images (one collection of images for $P$ and an independent collection for $Q$). The function space was parameterized via fully-connected neural networks with one hidden layer of 128 nodes. The results were averaged over 50 runs. 
  }\label{fig:Digits_shift}
\end{figure}

Next we illustrate the accelerated speed of convergence on  high-dimensional ($28\times 28=784$ dimensional)  realistic data by estimating the Hellinger divergence between two distributions obtained by (iid) randomly translating the MNIST handwritten digits image dataset \cite{MNIST}.  This provides an effective test case wherein we know the exact answer ($D_{f_{1/2}}(Q\|P)=0$). Figure \ref{fig:Digits_shift} shows the error, as a function of the number of SGD iterations, and once again demonstrates that the improved variational formulas lead to faster convergence; in this case, nearly one order of magnitude fewer SGD iterations are required to reach an  accuracy of $10^{-2}$ when using the tighter objective functionals. In practice, this means that one can more quickly  detect whether or not the two data streams are in fact coming from the same distribution.

\begin{figure}[ht]
\begin{minipage}[b]{0.49\linewidth}
\centering
\includegraphics[width=\textwidth]{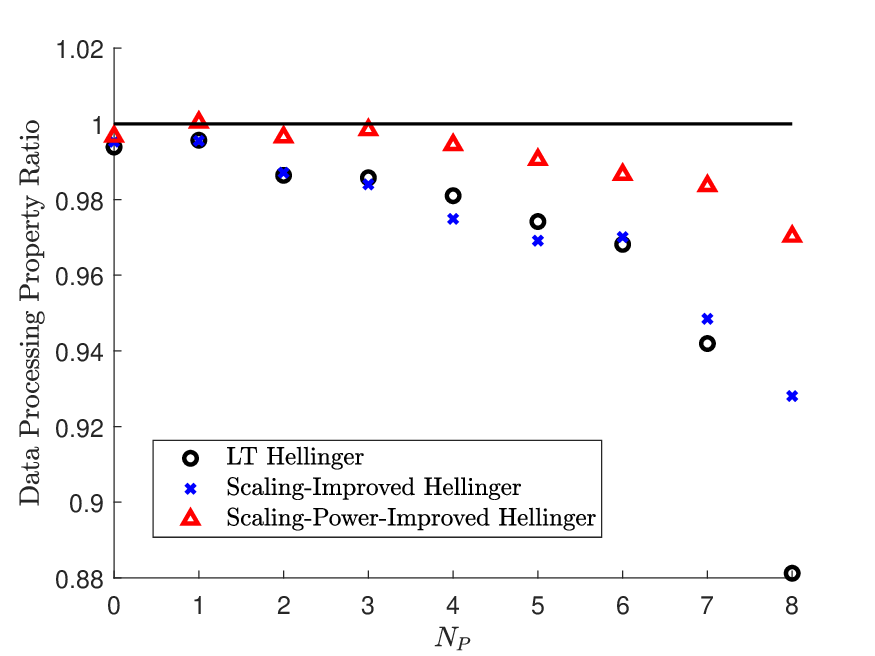}
\end{minipage}
\begin{minipage}[b]{0.49\linewidth}
\centering
\includegraphics[width=\textwidth]{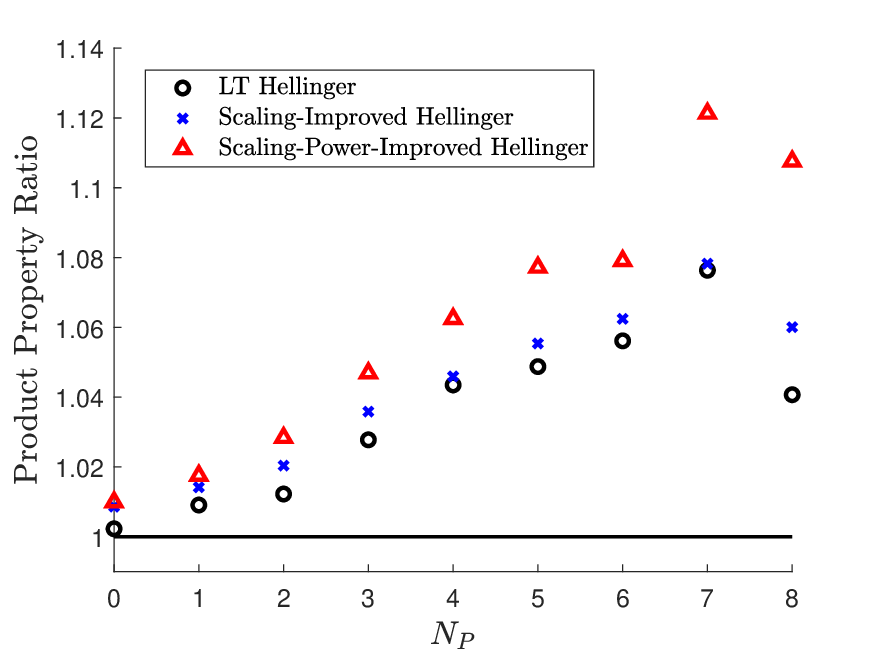}
\end{minipage}
\caption{{ 
Hellinger divergence consistency tests (Left: testing the data processing inequality \eqref{eq:data_proc}. Right: testing the product property \eqref{eq:alpha_additivity}) using MNIST dataset. We employ a fully connected neural network with one hidden layer of 128 nodes.  Training was performed with a minibatch size of 100.  We show results after 5000 steps of SGD and results were averaged over 100 runs.}}
  \label{fig:MNIST_consistency_checks}

\end{figure}

{

To further illustrate that our estimators are behaving appropriately, we perform a pair of consistency checks using the MNIST dataset, similar to the tests in \cite{Song2020Understanding}. While not a perfect substitute for computing the relative error, tests such as these are very useful in situations where   the exact value of the divergence is nonzero and unknown. In  Figure \ref{fig:MNIST_consistency_checks}(a) we test the data processing inequality for $f$-divergences:
\begin{align}\label{eq:data_proc}
 D_f(Q\otimes\kappa\| P\otimes \kappa)=D_f(Q\|P)\,,
\end{align}
where $\kappa$ is a probability kernel. Here we let $Q$ be the MNIST dataset, $P$ be the MNIST dataset of digits $0$ through $N_P$ where $N_P$ ranges from $0$ to $8$, and we let $\kappa_x$ be the distribution of random translations of the image $x$ (specifically, $N(0,1)$ translations, with components rounded to the nearest integer). The plot shows the ratio of the estimators for $D_{f_{1/2}}(Q\otimes\kappa\| P\otimes \kappa)$ and $D_{f_{1/2}}(Q\|P)$, using various objective functionals. In Figure \ref{fig:MNIST_consistency_checks}(b) we test the product measure identity for $\alpha$-divergences:
\begin{align}\label{eq:alpha_additivity}
& D_{f_\alpha}(Q_1\times...\times Q_k\| P_1\times...\times P_k)=\frac{\prod_{i=1}^k(\alpha(\alpha-1)D_{f_\alpha}(Q_i\|P_i)+1)-1}{\alpha(\alpha-1)}\,.
\end{align}
Here we let $Q_i$ be copies of the MNIST dataset and $P_i$ be copies of the MNIST dataset of digits $0$ through $N_P$, where $N_P$ again ranges from $0$ to $8$. The plot shows the ratio of the right-hand-side of \req{eq:alpha_additivity} to the left-hand-side. The black horizontal lines in both panels show the exact value of $1$. We find that all methods perform  well on these consistency checks, though the best performing method  is different for the two tests. This is unsurprising, as our methods are designed to accelerate the convergence to the optimum value, which is a different goal than preserving these the above two properties. Note that as $N_P$ increases the distributions $Q$ and $P$ become more similar and so both the numerator and denominator approach zero, making the task of estimating the ratio more difficult; this is reflected in Figure \ref{fig:MNIST_consistency_checks}.

}

To conclude these examples, we note that although the proposed optimization framework was applied on statistical learning and estimation, it can be of broader interest, among others,  in  epistemic uncertainty quantification \cite{chowdhary_dupuis_2013}, in coarse-graining and  model reduction \cite{Shell,Majda_fidelity,KP2013}, as well as in PAC learning \cite{catoni2008pac} and adversarial learning \cite{GAN, WGAN}. 
{
In particular, we intend to explore the use of variational formulas derived via the quadratic approximation method (e.g., \req{eq:approx_improved_DV}  and \req{eq:approx_improved_Renyi2}) for uncertainty quantification, along with the UQ bound \eqref{eq:UQ_Df}.
}


\section{Proof of Theorem \ref{thm:improved_variational_formula_general}}\label{sec:L1_variational_formula}

In this section we provide a detailed proof of Theorem \ref{thm:improved_variational_formula_general}.  For the convenience of the reader, we will recall the relevant definitions and notation below.

Let $P,Q$ be probability measures on a measurable space $(\Omega,\mathcal{M})$ and, for any $-\infty\leq a<1<b\leq \infty$ define $\mathcal{F}_1(a,b)$ to be the set of convex functions $f:(a,b)\to\mathbb{R}$ with $f(1)=0$. If $a$ (resp. $b$) is finite, we extend $f$ to $a$ (resp. $b$) by continuity and set $f(x)=\infty$ for $x\not\in[a,b]$.  The result is a convex, lower semicontinuous function, $f:\mathbb{R}\to(-\infty,\infty]$.  The $f$-divergence of $Q$ with respect to  $P$ is defined  by
\begin{align}
D_f(Q\|P)=\begin{cases} 
     E_P[f(dQ/dP)], & Q\ll P\\
      \infty, &Q\not\ll P
   \end{cases}\,.
\end{align}
Our starting point is the the following variational characterization  \cite{Broniatowski,Nguyen_Full_2010}:
\begin{align}\label{eq:Df_variational_bounded_app}
D_f(Q\|P)=\sup_{\phi\in\mathcal{M}_b(\Omega)}\{E_Q[\phi]-E_P[f^*(\phi)]\}\,,
\end{align}
where $\mathcal{M}_b(\Omega)$ denotes the set of all bounded and measurable functions and 
\begin{align}
   f^*(y)=\sup_{x\in\mathbb{R}}\{yx-f(x)\}=\sup_{x\in(a,b)}\{yx-f(x)\} 
\end{align} 
is the Legendre transform of $f$.
\begin{remark}
Note that $f^*(y)\geq y$. This implies $f^*(\phi)\geq \phi$  is bounded below for $\phi\in\mathcal{M}_b(\Omega)$, and hence $E_P[f^*(\phi)]\in (-\infty,\infty]$ is always well-defined.
\end{remark}

The technical aspects of the proof of  Theorem \ref{thm:improved_variational_formula_general} revolve around ensuring that all of the required expectations and operations are well-defined (without requiring any arbitrary convention regarding the definition of $\infty-\infty$).
Modulo those details, the derivation of \req{eq:Df_unbounded_3} is quite simple.  As a first step, we show that \req{eq:Df_variational_bounded_app}  can be extended to certain unbounded $\phi$. This is similar to results in \cite{Broniatowski,Nguyen_Full_2010} but we will prove explicit conditions for which the expectations exist. To do this, we will need the following lemmas:
\begin{lemma}\label{lemma:f_star_nondec}
Let   $f\in\mathcal{F}_1(a,b)$.  Then one of the following holds
\begin{enumerate}
\item $f^*$ is bounded below.
\item The set $\dom(f^*)\equiv\{y:f^*(y)<\infty\}$ is of the form $\dom(f^*)=(-\infty,d)$ or $\dom(f^*)=(-\infty,d]$ for some $d\in(-\infty,\infty]$ and $f^*$ is non-decreasing.
\end{enumerate}
\end{lemma}
\begin{IEEEproof}
Suppose $f^*$ is not bounded below.  Take $y_n\in I$ with $f^*(y_n)\to -\infty$.  We know $f^*(y)\geq y$ and so $y_n\leq f^*(y_n)\to -\infty$ and hence $y_n\to-\infty$.  $I$ is convex so it we let $d=\sup I$ then this implies  $(-\infty,d)\subset I\subset (-\infty,d]$. 

To show $f^*$ is non-decreasing, suppose  that we have $x_1<x_2$ with  $f^*(x_1)>f^*(x_2)$.  Taking $y_n$ as above, find an  $n$ such that $y_n<x_1$ and $f^*(y_n)<f^*(x_2)$.    $f^*$ is convex and so, letting $t=(x_2-x_1)/(x_2-y_n)\in(0,1)$, we have
\begin{align}
f^*(x_1)=&f^*(ty_n+(1-t)x_2)\leq tf^*(y_n)+(1-t)f^*(x_2)\notag\\
<&  tf^*(x_2)+(1-t)f^*(x_2)=f^*(x_2)<f^*(x_1)\,.
\end{align}
This is a  contradiction, hence $f^*$ is non-decreasing. 
\end{IEEEproof}
\begin{lemma}\label{lemma:f_star_int}
Let $f\in\mathcal{F}_1(a,b)$ and suppose $f^*$ is bounded below or $D_f(Q\|P)<\infty$. Then $E_P[f^*(\phi)^-]<\infty$ for all  $\phi\in L^1(Q)$.
\end{lemma}
\begin{remark}
We use the notation $g^+\equiv g1_{g\geq 0}$ and $g^-\equiv -g1_{g\leq 0}$ for  positive and negative parts of a function $g=g^+-g^-$.\end{remark}
\begin{IEEEproof}
Fix $\phi\in L^1(Q)$.  If $f^*$ is bounded below then $(f^*)^-$ is bounded above and the result is trivial, so suppose not.  Lemma \ref{lemma:f_star_nondec} then implies that   $f^*$ is non-decreasing. General properties of Legendre transforms on the real line imply that $f^*$ is convex, lower semicontinuous, and $f^*$ is continuous on $\overline{\dom(f^*)}$. Hence there exists $b\in\mathbb{R}$ such that  $f^*\leq 0$ on $(-\infty,b]$ and $f^*\geq 0$ on $(b,\infty)$ (note that  $f^*(0)\geq 0$).   Define
 $\phi_b=\phi1_{\phi\leq b}+ b1_{\phi> b}$, so that $\phi_b\leq b$, $\phi_b\in L^1(Q)$, and
\begin{align}
&E_P[f^*(\phi)^-]=E_P[1_{\phi\leq b}f^*(\phi)^-]=E_P[1_{\phi\leq b}f^*(\phi_b)^-]\leq E_P[f^*(\phi_b)^-]=E_P[-f^*(\phi_b)]\,.
\end{align}
Hence $E_P[f^*(\phi_b)]\leq -E_P[f^*(\phi)^-]$.

Now define $\phi_{b,n}=-n1_{\phi_b<-n}+\phi_b1_{\phi_b\geq -n}$.  $\phi_b$ is bounded above and so $\phi_{b,n}\in\mathcal{M}_b(\Omega)$ and we can use \req{eq:Df_variational_bounded_app} to find
\begin{align}
E_Q[\phi_{b,n}]\leq D_f(Q\|P)+E_P[f^*(\phi_{b,n})]\,.
\end{align}
We have $\phi_{b,n}\to\phi_b$ pointwise, $|\phi_{b,n}|\leq |\phi_b|$, and $\phi_b\in L^1(Q)$, so we can use the dominated convergence theorem to obtain 
\begin{align}
E_Q[\phi_{b}]\leq D_f(Q\|P)+\liminf_n E_P[f^*(\phi_{b,n})]
\end{align}
(here it was important that we are in the case where $D_f(Q\|P)<\infty$). We also have $\phi_{b,{n+1}}\leq  \phi_{b,n}$, hence $f^*( \phi_{b,{n+1}})\leq f^*( \phi_{b,n})$ (recall we are in the case where $f^*$ is nondecreasing) and for $N$ large enough we have $ \phi_{b,n}\leq b$ for all $n\geq N$.  $f^*$ is continuous on $(-\infty,b]$, hence so
\begin{align}
 0\leq-f^*( \phi_{b,n})\nearrow -  f^*( \phi_{b})\,.
\end{align}
Therefore the monotone convergence theorem implies $\lim_n E_P[f^*(\phi_{b,n})]= E_P[f^*(\phi_b)]$, and so
\begin{align}
&-\infty<E_Q[\phi_{b}]\leq D_f(Q\|P)+E_P[f^*(\phi_b)]\leq D_f(Q\|P) -E_P[f^*(\phi)^-]\,.
\end{align}
We therefore conclude that $E_P[f^*(\phi)^-]< \infty$. 
\end{IEEEproof}

We can now prove that \req{eq:Df_variational_bounded_app} can be extended to $\phi\in L^1(Q)$.
\begin{theorem}\label{thm:L1_optim}
Let $f\in\mathcal{F}_1(a,b)$ and suppose either $f^*$ is bounded below or $D_f(Q\|P)<\infty$. Then
\begin{align}
D_f(Q\|P)=&\sup_{\phi\in L^1(Q)}\{E_Q[\phi]-E_P[f^*(\phi)]\}\,,\label{eq:Df_unbounded_1_proof}
\end{align}
where the objective functional is valued in $[-\infty,\infty)$.
\end{theorem}
\begin{IEEEproof}
Lemma \ref{lemma:f_star_int} implies $E_P[f^*(\phi)^-]<\infty$ for all  $\phi\in L^1(Q)$ and so the objective functional in \req{eq:Df_unbounded_1_proof} is valued in $[-\infty,\infty)$.   If we can show $E_Q[\phi]-E_P[f^*(\phi)]\leq D_f(Q\|P)$ for all $\phi\in L^1(Q)$ then the claimed result will follow by using \req{eq:Df_variational_bounded_app}.

Fix $\phi\in L^1(Q)$.  If $D_f(Q\|P)=\infty$ or $E_P[f^*(\phi)]=\infty$ then the required bound is trivial, so suppose not.  Then $f^*(\phi)<\infty$ $P$-a.s.  We are in the case where $D_f(Q\|P)<\infty$, and so $Q\ll P$ and $f^*(\phi)<\infty$ $Q$-a.s. as well.  

In summary, it suffices to show $E_Q[\phi]-E_P[f^*(\phi)]\leq D_f(Q\|P)$ in the case where $\phi\in L^1(Q)$,  $f^*(\phi)\in L^1(P)$, $D_f(Q\|P)<\infty$, $\range(\phi)\subset \dom(f^*)$. To do this, fix $y_0\in I$ and define  $\phi_n=y_01_{\phi<-n}+\phi1_{-n\leq\phi\leq n}+y_01_{\phi>n}$.  $\phi_n\in\mathcal{M}_b(\Omega)$ and so \req{eq:Df_variational_bounded_app} gives
\begin{align}
D_f(Q\|P)\geq  E_Q[\phi_n]-E_P[f^*(\phi_n)]\,.
\end{align}
We have $\phi_n\to\phi$ pointwise and $|\phi_n|\leq |\phi|+|y_0|\in L^1(Q)$, therefore the dominated convergence theorem $E_Q[\phi_n]\to E_Q[\phi]$.

 We have $\range(\phi_n),\range(\phi)\subset \dom(f^*)$ and $f^*$ is continuous on $\dom(f^*)$, therefore $f^*(\phi_n)\to f^*(\phi)$ pointwise.  We also have
\begin{align}
|f^*(\phi_n)|=&|f^*(\phi_n)|1_{\phi<-n}+|f^*(\phi_n)|1_{-n\leq\phi\leq n}+|f^*(\phi_n)|1_{\phi>n}\\
\leq& |f^*(y_0)|+|f^*(\phi)|\in L^1(P)\,,\notag
\end{align}
hence the dominated convergence theorem implies $E_P[f^*(\phi_n)]\to E_P[f^*(\phi)]$.  Combining these gives
\begin{align}
E_Q[\phi_n]-E_P[f^*(\phi_n)]\to E_Q[\phi]- E_P[f^*(\phi)]
\end{align}
and so
\begin{align}
D_f(Q\|P)\geq  E_Q[\phi]- E_P[f^*(\phi)]\,.
\end{align}
This proves the claim.
\end{IEEEproof}

We now prove Theorem \ref{thm:improved_variational_formula_general} from the main text, which we restate below.  For completeness, we also provide a derivation of the formula for the optimizer, $\phi^*$, which was obtained in \cite{Broniatowski}.
\begin{theorem}\label{thm:Df_var_unbounded_L1Q}
Let $f\in\mathcal{F}_1(a,b)$ and suppose either $f^*$ is bounded below or $D_f(Q\|P)<\infty$. Then:
\begin{enumerate}
\item Suppose $Q\ll P$, $f$ is $C^1$, $f^\prime$ is strictly increasing, and one of the following holds:
\begin{enumerate}
    \item $a<dQ/dP<b$
    \item $a\leq dQ/dP\leq b$  and  if the value $a$ (resp. $b$) is achieved then $f^\prime(a)\equiv\lim_{t\searrow a} f^\prime(t)$ (resp. $f^\prime(b)\equiv\lim_{t\nearrow b}f^\prime(t)$) exists and is finite.
\end{enumerate}
Define $\phi^*=f^\prime(dQ/dP)$.  If $\phi^*\in L^1(Q)$ then the supremum in \req{eq:Df_unbounded_1_proof} is achieved at $\phi^*$.
\item Let $\Phi$ be a family of functions with $\mathcal{M}_b(\Omega)\subset\Phi\subset L^1(Q) \,\,\text{ or with }\,\,\phi^*\in\Phi\subset L^1(Q)$ (in the latter case, we also assume that the conditions from part (1) hold, so that $\phi^*$ is the optimizer). Consider any  family of transformations $\mathcal{T}\subset\{T=T(\phi)\, , \mbox{ such that }\,  T:\Phi \mapsto  L^1(Q)\}  $
that includes the identity map. Then
\begin{align}
D_f(Q\|P)=\sup_{\phi\in \Phi}\{\sup_{T\in\mathcal{T}}\{E_Q[T(\phi)]-E_P[f^*(T(\phi))]\}\} \, .\label{eq:Df_unbounded_proof}
\end{align}
\item If $Q\ll P$ and $(\Omega,\mathcal{M})$ is a metric space with the Borel $\sigma$-algebra then one can replace $\mathcal{M}_b(\Omega)$ with $C_b(\Omega)$ (bounded  continuous functions) and $L^1(Q)$ with $L^1_c(Q)$ (continuous  $L^1(Q)$ functions) in the above.
\end{enumerate}

\end{theorem}
\begin{remark}
Recall that $-\infty\leq a<1<b\leq \infty$. In particular, one could have $b=\infty$ and so the assumptions of part (1) do not necessary imply $dQ/dP$ is bounded.
\end{remark}
\begin{IEEEproof}
Suppose that $f$ satisfies the additional conditions from item (1).  Then the Legendre transform can be computed:
\begin{align}
    f^*(y)=y (f^\prime)^{-1}(y)-f((f^\prime)^{-1}(y))
\end{align}
for all $y\in\range(f^\prime)$. This implies that for $x\in (a,b)$ we have $f(x)=xf^\prime(x)- f^*(f^\prime(x))$.  By taking limits and using the assumptions on $dQ/dP$, we find $f(dQ/dP)=dQ/dP f^\prime(dQ/dP)-f^*(f^\prime(dQ/dP))$. Therefore, assuming $\phi^*\equiv f^\prime(dQ/dP)\in L^1(Q)$, we have
\begin{align}
&E_Q[\phi^*]-E_P[f^*(\phi^*)]\\
=&E_P[f^\prime(dQ/dP)dQ/dP -f^*(f^\prime(dQ/dP))]\notag\\
=&E_P[f(dQ/dP)]=D_f(Q\|P)\,,\notag
\end{align}
which proves the claim.

Now, let $\Phi$ and $\mathcal{T}$ be as in item (2).  Since every $T\in\mathcal{T}$ maps $\Phi$ into $L^1(Q)$ and $\mathcal{T}$ contains the identity we can use Theorem \ref{thm:L1_optim} to obtain
\begin{align}
 &   E_Q[\phi]-E_P[f^*(\phi)]\leq \sup_{T\in\mathcal{T}}\left\{ E_Q[T(\phi)]-E_P[f^*(T(\phi))]\right\}\leq D_f(Q\|P)
\end{align}
for all $\phi\in \Phi$.  Maximizing over $\Phi$ and using the fact that either $\phi^*\in L^1(Q)$ or $\mathcal{M}_b\subset \Phi\subset L^1(Q)$ along with \req{eq:Df_unbounded_1_proof} gives 
\begin{align}
&   D_f(Q\|P)\leq\sup_{\phi\in\Phi}\{ E_Q[\phi]-E_P[f^*(\phi)]\}\\
\leq& \sup_{\phi\in\Phi}\sup_{T\in\mathcal{T}}\left\{ E_Q[T(\phi)]-E_P[f^*(T(\phi))]\right\}\leq D_f(Q\|P)\,,\notag
\end{align}
which proves the claim.

Finally, on a metric space, one can approximate measurable functions with continuous functions via Lusin's theorem (see, e.g., Appendix D in \cite{dudley1999uniform}). Using this fact it is straightforward to show that the same results are obtained if one replaces $\mathcal{M}_b(\Omega)$ with $C_b(\Omega)$ and $L^1(Q)$ with $L^1_c(Q)$ (see the Proof of Theorem 1  in \cite{doi:10.1137/20M1368926} for details on the use of this technique in a similar context). This proves item (3).

\end{IEEEproof}

\appendices

\section{Derivation of Variational Formulas for $\alpha$ and R{\'e}nyi Divergences}\label{app:alpha_div} Here we provide additional details regarding the derivation of the new variational formulas for $\alpha$-divergences, as well as their connection to the R{\'e}nyi divergences.

\medskip
\noindent
\textbf{1.} 
Recall that the $\alpha$-divergences are the family of $f$-divergences corresponding to
    \begin{align}
        f_\alpha(t)=\frac{t^\alpha-1}{\alpha(\alpha-1)},\,\,\,\alpha\neq 0,1\,.
    \end{align}
   First consider the case $\alpha>1$:   The Legendre transform of $f_\alpha$ is 
 \begin{align}
f_\alpha^*(y)= y^{\alpha/(\alpha-1)}\alpha^{-1}(\alpha-1)^{\alpha/(\alpha-1)}+\frac{1}{\alpha(\alpha-1)},\,\,\,\,y\geq 0
\end{align}
and the exact optimizer \eqref{eq:optimum_f_divergence} is non-negative. With this in mind, fix $\phi\geq 0$, $\phi\in L^1(Q)$, and $\phi^{\alpha/(\alpha-1)}\in L^1(P)$, with $\phi$  not $P$-a.s. zero.  For $\eta>0$ we can write
\begin{align}
& E_Q[\eta\phi ] -E_P[f_\alpha^*(\eta\phi)]\\
=&\eta E_Q[\phi]-\eta^{\alpha/(\alpha-1)} \alpha^{-1}(\alpha-1)^{\alpha/(\alpha-1)}E_P[\phi^{\alpha/(\alpha-1)}]-\frac{1}{\alpha(\alpha-1)}\,.\notag
\end{align}
For $a,c\geq 0$, $b>0$ we have the following general solution to the optimization problem
\begin{align}
\sup_{\eta>0}\{a\eta-b\eta^{\alpha/(\alpha-1)}-c\}=\frac{a^\alpha}{\alpha b^{\alpha-1}}\left(\frac{\alpha-1}{\alpha}\right)^{\alpha-1}-c\,,
\end{align}
where the maximum occurs at  $\eta^*=(a(\alpha-1)/(b\alpha))^{\alpha-1}$. Hence we can use this formula to obtain
\begin{align}\label{eq:alpha_div_sup_eta}
& \sup_{\eta>0}\{E_Q[\eta\phi ] -E_P[f_\alpha^*(\eta\phi)]\}
=\frac{1}{\alpha  (\alpha-1)}E_Q[\phi]^\alpha E_P[\phi^{\alpha/(\alpha-1)}]^{-(\alpha-1)}-\frac{1}{\alpha(\alpha-1)}\,.
\end{align}
Using this, for $Q\ll P$ with $(dQ/dP)^\alpha\in L^1(P)$ we find
\begin{align}\label{eq:alpha_div_eta_opt_app}
&    D_{f_\alpha}(Q\|P=\frac{1}{\alpha(\alpha-1)}\sup_{\phi\in L^1(Q):\phi\geq 0}\{E_Q[\phi]^\alpha E_P[\phi^{\alpha/(\alpha-1)}]^{1-\alpha}-1\}\,,
\end{align}
and the maximum is achieved at
\begin{align}\label{eq:exact_opt_alpha_div}
    \phi_\alpha^*=(dQ/dP)^{\alpha-1}\,.
    \end{align}
In \req{eq:alpha_div_eta_opt_app} one should interpret $0/0\equiv 1$ and $c/\infty\equiv 0$, or else restrict to the subset of functions with $0<E_P[\phi^{\alpha/(\alpha-1)}]<\infty$ for which these cases do not occur. Also note that the objective functional  \eqref{eq:alpha_div_eta_opt_app} is now invariant under scaling and so we were able to drop the factor of $1/(\alpha-1)$ in \req{eq:exact_opt_alpha_div}.

If $\alpha\in(0,1)$ and $y<0$   then 
\begin{align}
   f_\alpha^*(y)= |y|^{-\alpha/(1-\alpha)}\alpha^{-1}(1-\alpha)^{-\alpha/(1-\alpha)}-\frac{1}{\alpha(1-\alpha)}\,,
\end{align}
and, if $Q\ll P$,  a similar computation to the above yields
\begin{align}\label{eq:Dfalpha_eta_imp_app}
&    D_{f_\alpha}(Q\|P)=\frac{1}{\alpha(1-\alpha)}\sup_{\phi>0}\left\{1-E_Q[\phi]^\alpha E_P[\phi^{-\alpha/(1-\alpha)}]^{1-\alpha}\right\}\,.
\end{align}
If $dQ/dP>0$ then the exact optimizer is $\phi^*_\alpha=(dQ/dP)^{\alpha-1}$. Note that we reparameterized $\phi\to-\phi$ in \req{eq:Dfalpha_eta_imp_app} so that we optimize over strictly positive functions, rather than strictly negative functions. Also, note that it is not necessary to restrict the optimization in \eqref{eq:Dfalpha_eta_imp_app} to $\phi\in L^1(Q)$; if $E_Q[\phi]=\infty$ then the objective functional  equals $-\infty$ and hence including such $\phi$'s does not change the value of the supremum. Theorem \ref{thm:improved_variational_formula_general} guarantees that the objective functionals in the variational representations \req{eq:alpha_div_eta_opt_app}, and \req{eq:Dfalpha_eta_imp_app} are tighter than that of the LT $f$-divergence objective functional from \eqref{eq:Df_variational_bounded_app}.

\medskip
\noindent
\textbf{2.}
In particular, for the Hellinger distance ($\alpha=1/2$) we find the scaling-improved formula
\begin{align}
  D_{f_{1/2}}(Q\|P)=4\sup_{\phi>0}\left\{1-E_Q[\phi]^{\frac{1}{2}} E_P[\phi^{-1}]^{\frac{1}{2}}\right\}\,,  
\end{align}
as compared to the LT $f$-divergence variational formula (after changing variables $\phi\to -\phi$)
\begin{align}
    D_{f_{1/2}}(Q\|P)=&\sup_{\phi>0}\{4-E_Q[\phi]-4E_P[\phi^{-1}]\}\,.
    \end{align}
    The improved tightness, $4-E_Q[\phi]-4E_P[\phi^{-1}] \leq 4(1-E_Q[\phi]^{\frac{1}{2}} E_P[\phi^{-1}]^{\frac{1}{2}})$, is guaranteed by the general result of Theorem \ref{thm:Df_var_unbounded_L1Q}, but it can also be seen to be a consequence of the inequality $ab\leq a^2+b^2/4$ applied to $a=E_P[\phi^{-1}]^{\frac{1}{2}}$, $b=E_Q[\phi]^{\frac{1}{2}}$.


\medskip
\noindent
\textbf{3.}
For the $\chi^2$-divergence (equal to $2D_{f_2}$) one can evaluate the optimization over all affine tranformations. We have $f_2(x)=(x^2-1)/2$ and $f_2^*(y)=(y^2+1)/2$.  Suppose $D_{f_2}(Q\|P)<\infty$.  Optimizing first over shifts we find, for $\phi\in L^1(Q)\cap L^2(P)$,
\begin{align}
&    \sup_{\nu\in\mathbb{R}}\{E_Q[\phi-\nu]-\frac{1}{2}E_P[(\phi-\nu)^2+1]\}\\
=&E_Q[\phi]+\frac{1}{2}(E_P[\phi]-1)^2-\frac{1}{2}(1+E_P[\phi^2])\notag\\
    =&E_Q[\phi]-E_P[\phi]-\frac{1}{2}\Var_P[\phi]\,,\notag
\end{align}
with the maximum occurring at $\nu^*=E_P[\phi]-1$. Therefore we obtain the variational representation
\begin{align}\label{eq:chi2_shift}
    \chi^2(Q\|P)=&\sup_{\phi\in L^1(Q)\cap L^2(P)}\{E_Q[2\phi]-E_P[2\phi]-\Var_P[\phi]\} \notag \\
    =&\sup_{\phi\in L^1(Q)\cap L^2(P)}\{E_Q[\phi]-E_P[\phi]-\frac{1}{4}\Var_P[\phi]\}\,.
\end{align}
Note that the objective functional in the last line is the same as that obtained in Eq. (48) of \cite{Bridging_fGan_WGan}.

Further optimizing the objective functional over the scaling parameter $\eta\in\mathbb{R}$ we find 
\begin{align}
    &\sup_{\eta\in\mathbb{R}}\{E_Q[\eta\phi]+\frac{1}{2}(E_P[\eta\phi]-1)^2-\frac{1}{2}(1+E_P[(\eta\phi)^2])\} \notag \\
  =&\frac{1}{2}\frac{(E_Q[\phi]-E_P[\phi])^2}{\Var_P[\phi]}
\end{align}
(if $\Var_P[\phi]>0$), with the maximum occurring at $\eta^*=(E_Q[\phi]-E_P[\phi])/\Var_P[\phi]$ (if $\Var_P[\phi]=0$ then the supremum equals zero, since $Q\ll P$).  Therefore
\begin{align}\label{eq:chi2_affine}
\chi^2(Q\|P)
=&2\sup_{\phi\in L^1(Q)}\sup_{\eta,\nu\in\mathbb{R}}\{E_Q[\eta\phi-\nu]-E_P[f^*_2(\eta\phi-\nu)]\}\\
=&\sup_{\phi\in L^1(Q)\cap L^2(P):\Var_P[\phi]>0}\frac{(E_Q[\phi]-E_P[\phi])^2}{\Var_P[\phi]}\,.\notag
\end{align}
Equality is achieved at $\phi^*=dQ/dP$ and so one can further restrict the optimization to $\phi\geq 0$.  This provides a rigorous justification of the  loss function for $\chi^2$-GANs proposed in, \cite{chi2_Gan}. We emphasize that the objective functional in \eqref{eq:chi2_affine}  is tighter than the one in \eqref{eq:chi2_shift}.

\medskip
\noindent
\textbf{4.}
The $\alpha$-divergences for $\alpha=0,1$ are the KL divergences: $D_{f_0}(Q\|P)=D_{KL}(P\|Q)$ and $D_{f_1}(Q\|P)=D_{KL}(Q\|P)$. Reparameterizing the objective functional in \req{eq:Dfalpha_eta_imp_app} via  
\begin{align}\label{eq:phi_Renyi_reparam}
\phi\to e^{\pm (\alpha-1)g}
\end{align}
(with the optimization running over all measurable $g$) provides  connections to the Donsker-Varadhan formula for the KL divergence in the limits $\alpha\to 0,1$.  Under the reparameterization  $\phi_\alpha=e^{ (\alpha-1)g}$ one has
\begin{align}
    E_Q[\phi_\alpha]^\alpha E_P[\phi_\alpha^{-\alpha/(1-\alpha)}]^{1-\alpha}
=&1+(E_Q[g]-\log E_P[e^g])(\alpha-1)+O((\alpha-1)^2)
\end{align}
and so
\begin{align}
    \lim_{\alpha\to 1}\frac{1}{\alpha(1-\alpha)}(1-E_Q[\phi_\alpha]^\alpha E_P[\phi_\alpha^{-\alpha/(1-\alpha)}]^{1-\alpha})  
=& E_Q[g]-\log E_P[e^g]\,,
\end{align}
which is the  the DV objective functional for $D_{KL}(Q\|P)$.

  Similarly, reparameterizing via $\phi_\alpha=e^{-(\alpha-1)g}$ gives
\begin{align}
E_Q[\phi_\alpha]^\alpha E_P[\phi_\alpha^{\alpha/(\alpha-1)}]^{-(\alpha-1)}
=&1-(E_P[g]-\log E_Q[e^g])\alpha +O(\alpha^2)
\end{align}
and so
\begin{align}
    \lim_{\alpha\to 0}\frac{1}{\alpha(1-\alpha)}(1-E_Q[\phi_\alpha]^\alpha E_P[\phi_\alpha^{-\alpha/(1-\alpha)}]^{1-\alpha})
     =& E_P[g]-\log E_Q[e^g]\,,
\end{align}
which is the objective functional for the Donsker-Varadhan representation of $D_{KL}(P\|Q)$.

\medskip
\noindent
\textbf{5.}
Using the same  reparametrization as in \req{eq:phi_Renyi_reparam} we can also derive a connection with the R{\'e}nyi family of divergences: Fix $\alpha\in(0,1)$, reparametrize   $\phi=e^{(\alpha-1)g}$ ($g$ is any measurable function), and rewrite \req{eq:Dfalpha_eta_imp_app} in terms of an optimization over $g$:
\begin{align}
    D_{f_\alpha}(Q\|P)
=&\frac{1}{\alpha(1-\alpha)}\sup_{g}\left\{1-E_Q[e^{(\alpha-1)g}]^\alpha E_P[e^{\alpha g}]^{1-\alpha}\right\}\,.
\end{align}
Using the  connection between $\alpha$-divergences and R{\'e}nyi divergences (see \cite{LieseVajda}, but note that our definition of R{\'e}nyi divergence differs from theirs by a factor of $1/\alpha$), we obtain
\begin{align}\label{eq:Renyi_var_SI}
    R_\alpha(Q\|P)
=&\frac{1}{\alpha(\alpha-1)}\log(\alpha(\alpha-1)D_{f_\alpha}+1)\\
    =&\frac{1}{\alpha(\alpha-1)}\log(\inf_{g}E_Q[e^{(\alpha-1)g}]^\alpha E_P[e^{\alpha g}]^{1-\alpha})\notag\\
    =& \sup_{g}\left\{\frac{1}{\alpha-1} \log( E_Q[e^{(\alpha-1)g}]) -\frac{1}{\alpha} \log( E_P[e^{\alpha g}])\right\}\,.\notag
\end{align}
A similar calculation applies when  $\alpha>1$; in either case, we obtain an independent derivation of the R{\'e}nyi divergence variational formula from \cite{doi:10.1137/20M1368926}.

\section{Improved Donsker-Varadhan Variational Formula}\label{app:Imp_DV} Here we collect some additional properties of the improved Donsker-Varadhan (DV) variational formula  \eqref{eq:improvedDV}:
    \begin{align}
D_{KL}(Q\|P)
=&\sup_{\phi\in \Phi}\{\sup_{\eta\in\mathbb{R}}\{\eta E_Q[\phi]-\log E_P[e^{\eta\phi}]\}\}\,.\label{eq:improvedDV_app}
\end{align}
One obviously has $\mathcal{T}^{{{id}}}\subset\mathcal{T}^{{shift}}\subset\mathcal{T}^{affine}$. As a consequence, the objective functional in \eqref{eq:improvedDV_app} is tighter than DV, which in turn is tighter than the LT $f$-Divergence objective functional from \req{eq:Df_variational_bounded_app}:
    \begin{align*}
   & \overbrace{\sup_{\eta\in\mathbb{R}} \{\eta E_Q[\phi]-\log E_P[e^{\eta\phi}]\}}^{\text{Improved DV \req{eq:improvedDV_app}}}\ge \overbrace{E_Q[\phi]-\log E_P[e^{\phi}]}^{\text{ DV \req{eq:DV}}}
\geq \overbrace{E_Q[\phi]-E_P[e^{\phi-1}]}^{\text{LT $f$-Divergence \req{eq:Df_variational_bounded_app}}}\,.
    \end{align*}

    Although the supremum over $\eta$ in \req{eq:improvedDV_app} cannot in general be evaluated analytically, one can obtain an explicit approximation as follows:     Define $G_\phi(\eta)=\eta E_Q[\phi]-\log E_P[e^{\eta\phi}]$ and Taylor expand around $\eta=1$ to obtain
\begin{align}\label{eq:G_taylor}
G_\phi(1+\Delta\eta)=&E_Q[\phi]-\log E_P[e^\phi]+(E_Q[\phi]-E_{P_\phi}[\phi])\Delta\eta-\frac{\Var_{P_\phi}[\phi]}{2}\Delta\eta^2+O(\Delta\eta^3)\,. 
\end{align}
We  approximate the optimal $\Delta \eta$ by maximizing the quadratic approximation \eqref{eq:G_taylor} to find
\begin{align}\label{eq:Delta_eta_star2}
\Delta\eta^*(\phi)=\frac{E_Q[\phi]-E_{P_\phi}[\phi]}{\Var_{P_\phi}[\phi]}\,,
\end{align}
where $dP_\phi=e^\phi dP/E_P[e^\phi]$ is the tilted measure and $\Delta\eta^*(\phi)$ is defined to be $0$ if $\Var_{P_\phi}[\phi]=0$. Using this,  we obtain a new variational representation:
\begin{theorem} \label{approx:improved:kld:thm} Define $\Phi=\{\phi\in L^1(Q):e^\phi,\phi e^\phi\in L^1(P)\}$ and for $\phi\in \Phi$ define $\Delta\eta^*(\phi)$ as in \req{eq:Delta_eta_star2}. Suppose $\log(dQ/dP)\in L^1(Q)$.  Then
\begin{align}\label{eq:approx_improved_DV2}
   D_{KL}(Q\|P)
=&\sup_{\phi\in \Phi}\{(1+\Delta\eta^*(\phi))E_Q[\phi]-\log E_P[e^{(1+\Delta\eta^*(\phi))\phi}]\}\,.
\end{align}
\end{theorem}
\begin{IEEEproof}
It is easy to see  that the objective functional in \req{eq:approx_improved_DV} is well defined for all $\phi\in\Phi$. To see that \req{eq:approx_improved_DV} is an equality, first recall that $D_{KL}(Q\|P)\geq G_\phi(\eta)$ for all $\eta$ and all $\phi\in L^1(Q)$ (see \req{eq:improvedDV_app}), and so $D_{KL}(Q\|P)\geq \sup_{\phi\in\Phi} G_\phi(1+\Delta \eta^*(\phi))$.  Computing the value at  $\phi^*=\log(dQ/dP)\in \Phi$ we see that $P_{\phi^*}=Q$ and so $\Delta\eta^*(\phi^*)=0$ and
\begin{align}
&\sup_{\phi\in \Phi}\{(1+\Delta\eta^*(\phi))E_Q[\phi]-\log E_P[e^{(1+\Delta\eta^*(\phi))\phi}]\}\\
\geq & (1+\Delta\eta^*(\phi^*))E_Q[\phi^*]-\log E_P[e^{(1+\Delta\eta^*(\phi^*))\phi^*}]\notag\\
=&E_Q[\phi^*]-\log E_P[e^{\phi^*}]=D_{KL}(Q\|P)\,.\notag
\end{align}
This proves  \eqref{eq:approx_improved_DV}.
\end{IEEEproof}

\req{eq:approx_improved_DV} is a new variational representation of the KL-divergence, however we make no claim that the  objective functional in \eqref{eq:approx_improved_DV} is  tighter than Donsker-Varadhan for every $\phi$; \req{eq:approx_improved_DV} is only guaranteed to be tighter than DV when $\Delta\eta^*(\phi)$ is sufficiently small. This is because we only maximized  the quadratic approximation in $\eta$, and hence only obtained an approximation to the scaling-improved objective functional from \req{eq:improvedDV_app}. One can of course circumvent this by using the maximum of the two:
\begin{align}\label{eq:DV_improved_max}    D_{KL}(Q\|P)
=\sup_{\phi\in\Phi}\bigg\{\max\big\{(1+\Delta\eta^*(\phi))E_Q[\phi]-\log E_P[e^{(1+\Delta\eta^*(\phi))\phi}],E_Q[\phi]-\log E_P[e^{\phi}]\big\}\bigg\}.
\end{align}
Although \req{eq:DV_improved_max} is less than aesthetically appealing, note that its objective functional is certainly no worse than DV and will be tighter when $\Delta\eta^*(\phi)$ is sufficiently small. Also note that no additional expectations need to be computed to evaluate the objective functional in \req{eq:DV_improved_max}    as compared to that of \req{eq:approx_improved_DV}.

{ 
\section{Approximating the Power-Improved R{\'e}nyi Variational Representation}
\label{app:Imp_Renyi}
The same method used to derive the second-order approximation of the improved Donsker-Varadhan variational formula \eqref{eq:approx_improved_DV} can be applied to the R{\'e}nyi divergence with power transformations. Recall that the power-improved  R{\'e}nyi variational is given by
\begin{align}\label{eq:Renyi_optim:power2}
    R_\alpha(Q\|P)
 =& \sup_{g} \sup_{\beta} \left\{\frac{1}{\alpha-1} \log( E_Q[e^{(\alpha-1)\beta g}]) -\frac{1}{\alpha} \log( E_P[e^{\alpha\beta g}])\right\}\, .
\end{align}
Define $G_{g,\alpha}(\beta) = \frac{1}{\alpha-1} \log( E_Q[e^{(\alpha-1)\beta g}]) -\frac{1}{\alpha} \log( E_P[e^{\alpha\beta g}])$ to be the objective functional as a function of $\beta$. A Taylor expansion around $\beta=1$ gives
\begin{align}
G_{g,\alpha}(1+\Delta\beta) &= G_{g,\alpha}(1) + (E_{Q_{\alpha-1}}[g]-E_{P_\alpha}[g]) \Delta\beta \\
&- \frac{1}{2} ((1-\alpha)\Var_{Q_{\alpha-1}}[g] + \alpha\Var_{P_\alpha}[g]) \Delta\beta^2 + O(\Delta\beta^3) \, , \notag
\end{align}
with $dQ_\alpha= e^{\alpha g} dQ/E_Q[e^{\alpha g}]$ being the $\alpha$-tilted measure and similarly for $P$. 
The quadratically-optimal $\Delta \beta^*$ is obtained by maximizing the second-order Taylor approximation and is given by
\begin{align}\label{eq:Delta_beta_star2}
\Delta\beta^*(g) = 
\frac{E_{Q_{\alpha-1}}[g]-E_{P_\alpha}[g]}{(1-\alpha)\Var_{Q_{\alpha-1}}[g] + \alpha\Var_{P_\alpha}[g]} \,,
\end{align}
where we assumed $0<\alpha < 1$. If $(1-\alpha)\Var_{Q_{\alpha-1}}[g] + \alpha\Var_{P_\alpha}[g]=0$ then we set $\Delta\beta^*(g) = 0$. A new R{\'e}nyi representation formula is then obtained as the following theorem asserts.
\begin{theorem}
\label{approx:power:renyi:thm}
Define $\Phi=\{g : g^ke^{(\alpha-1)g}\in L^1(Q), g^ke^{\alpha g}\in L^1(P) \ \text{with} \ k=0,1\}$ and for $g\in \Phi$ define $\Delta\beta^*(g)$ as in \req{eq:Delta_beta_star2}. Suppose $\log(dQ/dP)\in \Phi$.  Then for $\alpha\in(0,1)$ we have
\begin{align}\label{eq:approx_improved_Renyi2}
  R_\alpha(Q\|P) = &\sup_{g\in\Phi} \left\{\frac{1}{\alpha-1} \log( E_Q[e^{(\alpha-1) (1+\Delta\beta^*(g)) g}])-\frac{1}{\alpha} \log( E_P[e^{\alpha(1+\Delta\beta^*(g)) g}])\right\}\, .
\end{align}
\end{theorem}
\begin{IEEEproof}
The proof is similar to the proof of Theorem \ref{approx:improved:kld:thm}. First, the integrability assumptions ensure that $\Delta \beta^*(g)$, and hence the objective functional, are well-defined (the latter possibly equaling $-\infty$). Second, $R_\alpha(Q\|P)\geq G_{g,\alpha}(\beta)$ for all $\beta$ and $g\in\Phi$ thus $R_\alpha(Q\|P)\geq G_{g,\alpha}(1+\Delta\beta^*(g))$. It remains to show that the there is a $g$ such that the supremum is attained. Taking $g^*=\log(dQ/dP)\in \Phi$, it is sufficient to show that $\Delta\beta^*(g^*)=0$. We compute the two terms of the numerator of $\Delta\beta^*(g^*)$:
\begin{align}
E_{Q_{\alpha-1}}[g^*] =& \frac{E_Q[e^{(\alpha-1)\log(dQ/dP)}\log(dQ/dP)]}{E_Q[e^{(\alpha-1)\log(dQ/dP)}]} \notag \\
=& \frac{E_Q[(dQ/dP)^{(\alpha-1)}\log(dQ/dP)]}{E_Q[(dQ/dP)^{(\alpha-1)}]}\notag\\
=& \frac{E_P[(dQ/dP)^{\alpha}\log(dQ/dP)]}{E_P[(dQ/dP)^{\alpha}]} \, , \notag
\end{align}
and
\begin{align}
E_{P_\alpha}[g^*] =& \frac{E_P[e^{\alpha\log(dQ/dP)}\log(dQ/dP)]}{E_P[e^{\alpha\log(dQ/dP)}]}\\
= &\frac{E_P[(dQ/dP)^{\alpha}\log(dQ/dP)]}{E_P[(dQ/dP)^{\alpha}]}\, . \notag
\end{align}
Thus $\Delta\beta^*(g^*)=0$, which completes the proof.
\end{IEEEproof}

}

\section{Tightness Gains and Variational Derivatives}\label{app:Hessian}

{ Here we prove an extension of Theorem~\ref{thm:Hessian_KL} for the more general case of $f$-divergences. 

\begin{theorem}[Tightness gains for $f$-divergences]
\label{thm:Hessian_f_div}
In addition to the assumptions of Theorem~\ref{thm:improved_variational_formula_general} suppose $f^\prime$ is strictly increasing and $f^*\in C^3((c,d))$ with $(f^*)^{\prime\prime}>0$. We select  
the function space $\Phi=\mathcal{M}_b(\Omega)$ and assume that the maximizer $\phi^*$ in  \eqref{eq:optimum_f_divergence} is valued in a compact subset of $(c,d)$.
Then,  
for  $\mathcal{T}= \mathcal{T}^{{id}}\, ,
   \mathcal{T}^{{shift}}\, , \mathcal{T}^{{scale}}\, ,
   \mathcal{T}^{affine}$
we have that
$\mathcal{J}_{\mathcal{T}}$ in \eqref{eq:H_functional_epsilon} is twice differentiable 
at $\epsilon=0$ for any $\psi \in \mathcal{M}_b(\Omega)$. Furthermore, using the notation 
\eqref{eq:Hessian_def},
the corresponding 2nd Gateaux derivatives $\frac{d^2}{d\epsilon^2}\mathcal{J}_{\mathcal{T}}(0)=
\frac{d^2}{d\epsilon^2}\big|_{\epsilon=0}
H_{\mathcal{T}}[\phi^*+\epsilon \psi]$ 
for all   $\mathcal{T}= \mathcal{T}^{{id}}\, , \mathcal{T}^{{shift}}\, , \mathcal{T}^{affine}$ satisfy the following:
\begin{align}
  \langle\nabla^2H_{{id}}[\phi^*]\psi,\psi\rangle 
=&  -E_P[(f^*)^{\prime\prime}(\phi^*)]E_{P^*}[(\psi)^2]\notag\\
   =&   -E_P[(f^*)^{\prime\prime}(\phi^*)]\left(\Var_{P^*}[\psi]+E_{P^*}[\psi]^2\right)   \,, \label{H_unoptimized_f}\\
   \langle \nabla^2 H_{{shift}}[\phi^*]\psi,\psi\rangle=&-E_P[(f^*)^{\prime\prime}(\phi^*)]\Var_{P^*}[\psi]\,,
    \label{eq:hessian:gain:shift_f}\\
  \langle \nabla^2 H_{{scale}}[\phi^*]\psi,\psi\rangle =&  -E_P[(f^*)^{\prime\prime}(\phi^*)]\left(\Var_{P^*}[\psi]+E_{P^*}[\psi]^2-\frac{\left(E_{P^*}[\phi^*\psi]\right)^2}{E_{P^*}[(\phi^*)^2]}\right)\notag \\
 =&  -E_P[(f^*)^{\prime\prime}(\phi^*)]\left(E_{P^*}[(\psi)^2]-\frac{\left(E_{P^*}[\phi^*\psi]\right)^2}{E_{P^*}[(\phi^*)^2]}\right) 
  \,,\label{eq:hessian:gain:scalar_f}
   \\  
\langle \nabla^2H_{affine}[\phi^*]\psi, \psi\rangle=&-E_P[(f^*)^{\prime\prime}(\phi^*)]\left[
\Var_{P^*}[\psi] - \frac{\cov_{P^*}(\phi^*,\psi)^2}{\Var_{P^*}[\phi^*]}\right]\notag\\
=&-E_P[(f^*)^{\prime\prime}(\phi^*)]\Var_{P^*}[\psi]\left[1-\rho_{P^*}^2(\phi^*,\psi)\right]
\,,
\label{eq:hessian:gain:affine_f}
\end{align}
corresponding to \eqref{eq:Df_variational_bounded}, \eqref{eq:nu_improved},  \eqref{eq:eta_improved} and the combination of the last two respectively. The tilted probability measure $P^*$ is defined as 
\begin{align}\label{eq:P_star_def_thm}
    dP^*=(f^*)^{\prime\prime}(\phi^*)dP/E_P[(f^*)^{\prime\prime}(\phi^*)]\, ,
\end{align}
while $\rho_{P^*}(\phi^*, \psi)$ denotes the correlation between $\phi^*$ and $\psi$.
\end{theorem}
\begin{remark} 
The relations \eqref{H_unoptimized_f}, \eqref{eq:hessian:gain:shift_f}, \eqref{eq:hessian:gain:scalar_f}
and \eqref{eq:hessian:gain:affine_f} imply the following comparisons  in variational curvatures around the optimizer $\phi^*$, extending the discussion in Section~\ref{sec:Hessians} to $f$-divergences:
\begin{align}
&\langle \nabla^2H_{affine}[\phi^*]\psi, \psi\rangle 
\ge \langle \nabla^2 H_{{scale}}[\phi^*]\psi,\psi\rangle\ge \langle\nabla^2H_{{id}}[\phi^*]\psi,\psi\rangle
\end{align}
and 
\begin{align}
&\langle \nabla^2H_{affine}[\phi^*]\psi, \psi\rangle \ge \langle \nabla^2 H_{{shift}}[\phi^*]\psi,\psi\rangle\ge \langle\nabla^2H_{{id}}[\phi^*]\psi,\psi\rangle\,.
\end{align}
Furthermore, relations \eqref{H_unoptimized_f}, \eqref{eq:hessian:gain:shift_f}, \eqref{eq:hessian:gain:scalar_f}
and \eqref{eq:hessian:gain:affine_f} quantify precisely the  gains in each inequality; compare also to the demonstration in Figure~\ref{fig:Hessian_test}. We note that the inequality $\langle \nabla^2H_{affine}[\phi^*]\psi, \psi\rangle \ge \langle \nabla^2 H_{{scale}}[\phi^*]\psi,\psi\rangle$ readily follows from \eqref{eq:hessian:gain:scalar_f}
and \eqref{eq:hessian:gain:affine_f} when all pertinent terms are rewritten using the correlation $\rho_{P^*}(\phi^*, \psi)$.
\end{remark}
\begin{remark} In the KL case we readily have $P^*=Q$ and $E_P[(f^*)^{\prime\prime}(\phi^*)]=1$, thus Theorem~\ref{thm:Hessian_f_div} implies  the results of Theorem~\ref{thm:Hessian_KL}. 
\end{remark}
}

\begin{IEEEproof}For the convenience of the reader, we start by repeating  several of the equations from the main text. First define
\begin{align}
    H[\phi,\eta,\nu]=E_Q[\eta\phi-\nu]-E_P[f^*(\eta\phi-\nu)]\,.
\end{align}
Optimization over the affine family leads to four different variational representations of the $f$-divergence
 \begin{align}\label{eq:Df_H_app}
D_f(Q\|P)=&\sup_\phi \overbrace{H[\phi,1,0]}^{H_{{id}}[\phi]}=\sup_{\phi}\overbrace{\sup_{\nu}H[\phi,1,\nu]}^{H_{{shift}}[\phi]}\\
 =&\sup_{\phi}\overbrace{\sup_{\eta}H[\phi,\eta,0]}^{H_{{scale}}[\phi]} =\sup_{\phi}\overbrace{\sup_{\eta,\nu}H[\phi,\eta,\nu] }^{H_{affine}[\phi]}\,.\notag
\end{align} 
{ Then, using the notation 
\eqref{eq:Hessian_def} for the 2nd Gateaux derivatives, we have
the corresponding variational Hessians}
\begin{align}\label{eq:affine_hessians}
&
\langle \nabla^2H_{{id}}[\phi^*]\psi, \psi\rangle=\frac{d^2}{d\epsilon^2} \big|_{\epsilon=0}H[\phi^*+\epsilon\psi,1,0]\,,\\
&
\langle \nabla^2H_{{shift}}[\phi^*]\psi, \psi\rangle=\frac{d^2}{d\epsilon^2}\big|_{\epsilon=0} \sup_{\nu}H[\phi^*+\epsilon\psi,1,\nu]\,,\notag\\
&
\langle\nabla^2H_{{scale}}[\phi^*]\psi, \psi\rangle=\frac{d^2}{d\epsilon^2}\big|_{\epsilon=0} \sup_{\eta}H[\phi^*+\epsilon\psi,\eta,0]\,,\notag\\
&
\langle\nabla^2H_{affine}[\phi^*]\psi, \psi\rangle=\frac{d^2}{d\epsilon^2}\big|_{\epsilon=0} \sup_{\eta,\nu}H[\phi^*+\epsilon\psi,\eta,\nu]\,.\notag
\end{align}

The following general computation will facilitate the computation of the variational Hessians that appear in \req{eq:affine_hessians}: The maxima in \req{eq:Df_H_app} are achieved at $\eta=1$, $\nu=0$, $\phi=\phi^*\equiv f^\prime(dQ/dP)$. Given $C^2$ families $\phi_\epsilon=\phi^*+\epsilon\psi$, $\eta_\epsilon$ with $\eta_0=1$, and $\nu_\epsilon$ with $\nu_0=0$,  we can compute the derivatives
    \begin{align}
        &\frac{d}{d\epsilon}H[\phi_\epsilon,\eta_\epsilon,\nu_\epsilon]=E_Q[\eta_\epsilon^\prime\phi_\epsilon+\eta_\epsilon\psi-\nu_\epsilon^\prime]\\
&-E_P[(f^*)^\prime(\eta_\epsilon\phi_\epsilon-\nu_\epsilon)(\eta_\epsilon^\prime\phi_\epsilon+\eta_\epsilon\psi-\nu_\epsilon^\prime)]\,,\notag\\
&\frac{d^2}{d\epsilon^2}H[\phi_\epsilon,\eta_\epsilon,\nu_\epsilon]=E_Q[\eta_\epsilon^{\prime\prime}\phi_\epsilon+2\eta_\epsilon^\prime\psi-\nu_\epsilon^{\prime\prime}]\notag\\
        &-E_P[(f^*)^{\prime\prime}(\eta_\epsilon\phi_\epsilon-\nu_\epsilon)(\eta_\epsilon^\prime\phi_\epsilon+\eta_\epsilon\psi-\nu_\epsilon^\prime)^2]\notag\\
&-E_P[(f^*)^\prime(\eta_\epsilon\phi_\epsilon-\nu_\epsilon)(\eta_\epsilon^{\prime\prime}\phi_\epsilon+2\eta_\epsilon^\prime\psi-\nu_\epsilon^{\prime\prime})]\,.\notag
    \end{align}
    { Here we used the boundedness assumptions on $f^*$, the differentiablity of $f^*$ and the dominated convergence theorem  \cite{Folland} (Theorem 2.27); the latter allowed us to rigorously   exchange derivatives and expectations when we calculated the 1st and the 2nd Gateaux derivatives of the objective functionals above. The same argument is also used in the evaluation of all Gateaux derivatives below.}
    Furthermore, for any choice of $\psi$, $\nu_\epsilon$, $\eta_\epsilon$ that have the above properties, the maximum of $\epsilon\to H[\phi_\epsilon,\eta_\epsilon,\nu_\epsilon]$ is achieved at $\epsilon=0$.  Therefore the first derivative vanishes at $\epsilon=0$.  In particular, by considering the  case $\eta_\epsilon\equiv 1$, $\nu_\epsilon\equiv 0$ we see that
    \begin{align}\label{eq:H_first_deriv_app}
        E_P[(f^*)^\prime(\phi^*)\psi]=E_Q[\psi]
    \end{align}
    for all $\psi$.  The second derivative at the maximum is given by
\begin{align}
    &\frac{d^2}{d\epsilon^2}\big|_{\epsilon=0}H[\phi_\epsilon,\eta_\epsilon,\nu_\epsilon]\\
    =&E_Q[\eta_0^{\prime\prime}\phi^*+2\eta_0^\prime\psi-\nu_0^{\prime\prime}]-E_{P}[(f^*)^{\prime\prime}(\phi^*)(\eta_0^\prime\phi^*+\psi-\nu_0^\prime)^2]\notag\\
&-E_P[(f^*)^\prime(\phi^*)(\eta_0^{\prime\prime}\phi^*+2\eta_0^\prime\psi-\nu_0^{\prime\prime})]\notag\\
    =&-E_{P}[(f^*)^{\prime\prime}(\phi^*)(\eta_0^\prime\phi^*+\psi-\nu_0^\prime)^2]\,,\notag
\end{align}
where we used \req{eq:H_first_deriv_app} to cancel several terms. We will specialize this to compute all four  Hessians from \req{eq:affine_hessians}.

\subsection{ LT  $f$ - Divergence objective functional} 
Here we fix $\eta_\epsilon\equiv 1$, $\nu_\epsilon\equiv 0$:
\begin{align}
    H[\phi,1,0]=E_Q[\phi]-E_P[f^*(\phi)]
\end{align}
and hence
\begin{align}
    \langle \nabla^2H_{{id}}[\phi^*]\psi, \psi\rangle=&\frac{d^2}{d\epsilon^2}\big|_{\epsilon=0} H[\phi_\epsilon,1,0]\\
=& \frac{d^2}{d\epsilon^2}\big|_{\epsilon=0}(E_Q[\phi_\epsilon]-E_P[f^*(\phi_\epsilon)])\notag\\
    =&-E_P[(f^*)^{\prime\prime}(\phi^*)\psi^2]\,.\notag
\end{align}
We have assumed $(f^*)^{\prime\prime}>0$ and so we can define the probability measure 
\begin{align}\label{eq:P_star_def}
    dP^*=(f^*)^{\prime\prime}(\phi^*)dP/E_P[(f^*)^{\prime\prime}(\phi^*)]
\end{align}
and thereby write \begin{align}\label{eq:Hessian_eta1_nu0}
    \langle \nabla^2H_{{id}}[\phi^*]\psi, \psi\rangle=-E_P[(f^*)^{\prime\prime}(\phi^*)]E_{P^*}[\psi^2]
\end{align}
(compare with the KL-case \req{H_unoptimized}).

\subsection{ Optimization over shifts} Here $\eta_\epsilon\equiv 1$ and $\nu_\epsilon$ is chosen so that 
\begin{align}\label{eq:nu_eps_def}
    \sup_\nu\{-\nu-E_P[f^*(\phi_\epsilon-\nu)]\}=-\nu_\epsilon-E_P[f^*(\phi_\epsilon-\nu_\epsilon)]\,.
\end{align}
{ By using the convexity of the objective functional in $\nu$ along with implicit function theorem, one can see that the assumptions on $f$ are sufficient to ensure that such a smooth $\nu_\epsilon$ exists for  $\epsilon$ in a neighborhood of $0$.} We can simplify
\begin{align}
 &  \langle \nabla^2H_{{shift}}[\phi^*]\psi, \psi\rangle= \frac{d^2}{d\epsilon^2}\big|_{\epsilon=0}\sup_\nu H[\phi_\epsilon,1,\nu]\\
=&\frac{d^2}{d\epsilon^2}\big|_{\epsilon=0}H[\phi_\epsilon,\eta_\epsilon,\nu_\epsilon]=-E_{P}[(f^*)^{\prime\prime}(\phi^*)(\psi-\nu_0^\prime)^2]\,.\notag
\end{align}
The derivative of $\nu_\epsilon$ can be computed as follows:
\req{eq:nu_eps_def} implies 
\begin{align}
    0=&\partial_\nu|_{\nu=\nu_\epsilon}(-\nu-E_P[f^*(\phi_\epsilon-\nu)])=-1+E_P[(f^*)^\prime(\phi_\epsilon-\nu_\epsilon)]
\end{align}
for all $\epsilon$. Differentiating with respect to $\epsilon$ gives
\begin{align}
    0=E_P[(f^*)^{\prime\prime}(\phi_\epsilon-\nu_\epsilon)(\psi-\nu_\epsilon^\prime)]\,,
\end{align}
hence
\begin{align}
    \nu_0^\prime=\frac{E_P[(f^*)^{\prime\prime}(\phi^*)\psi]}{E_P[(f^*)^{\prime\prime}(\phi^*)]}=E_{P^*}[\psi]
\end{align}
where $P^*$ is the probability measure defined in \req{eq:P_star_def}.  Therefore
\begin{align}
    \frac{d^2}{d\epsilon^2}\big|_{\epsilon=0}\sup_\nu H[\phi_\epsilon,1,\nu]=&-E_P[(f^*)^{\prime\prime}(\phi^*)]E_{P^*}[ (\psi-\nu_0^\prime)^2]\notag\\
=&-E_P[(f^*)^{\prime\prime}(\phi^*)]\Var_{P^*}[\psi]
    \end{align}
(compare with the KL-case \req{eq:hessian:gain:shift}).

Recalling the result   \eqref{eq:Hessian_eta1_nu0} we arrive at
\begin{align}
   \langle \nabla^2H_{{shift}}[\phi^*]\psi, \psi\rangle=&\langle \nabla^2H_{{id}}[\phi^*]\psi, \psi\rangle+E_P[ (f^*)^{\prime\prime}(\phi^*)\psi]^2/E_P[(f^*)^{\prime\prime}(\phi^*)]\,.
\end{align}
This shows that the magnitude of the second derivative (which is guaranteed to be non-positive) has been reduced by the amount $E_P[ (f^*)^{\prime\prime}(\phi^*)\psi]^2/E_P[(f^*)^{\prime\prime}(\phi^*)]$.

\subsection{Optimization over scaling transformations} Here $\nu_\epsilon\equiv 0$ and $\eta_\epsilon$ is chosen so that
\begin{align}
    \sup_{\eta}\{E_Q[\eta\phi_\epsilon]-E_P[f^*(\eta\phi_\epsilon)]\}=E_Q[\eta_\epsilon\phi_\epsilon]-E_P[f^*(\eta_\epsilon\phi_\epsilon)]\,.
\end{align}
{ Again, the existence and smoothness of $\eta_\epsilon$ is guaranteed by convexity and the implicit function theorem.} Next note that
\begin{align}
 0= & \partial_\eta|_{\eta=\eta_\epsilon}(E_Q[\eta\phi_\epsilon]-E_P[f^*(\eta\phi_\epsilon)])
=E_Q[\phi_\epsilon]-E_P[(f^*)^\prime(\eta_\epsilon\phi_\epsilon)\phi_\epsilon]
\end{align}
for all $\epsilon$. Taking the derivative with respect to $\epsilon$ gives
\begin{align}
 0=  E_Q[\psi]-E_P[(f^*)^{\prime\prime}( \phi^*)(\eta_0^\prime\phi^*+\psi)\phi^*+(f^*)^\prime( \phi^*)\psi]\,.
 \end{align}
Solving for $\eta_0^\prime$ and using \req{eq:H_first_deriv_app} to simplify yields
\begin{align}
 \eta_0^\prime=&  -\frac{  E_P[(f^*)^{\prime\prime}(\phi^*)\psi\phi^*]}{E_P[(f^*)^{\prime\prime} (\phi^*)(\phi^*)^2]}\,.
 \end{align}
 Therefore, the Hessian is
\begin{align}
    &\langle \nabla^2H_{{scale}}[\phi^*]\psi, \psi\rangle=\frac{d^2}{d\epsilon^2}\big|_{\epsilon=0}\sup_{\eta}H[\phi_\epsilon,\eta,0] \\
    =&-E_{P}[(f^*)^{\prime\prime}(\phi^*)(\eta_0^\prime\phi^*+\psi)^2]\notag\\
    =&\langle \nabla^2H_{{id}}[\phi^*]\psi, \psi\rangle-(\eta_0^\prime)^2E_{P}[(f^*)^{\prime\prime}(\phi^*)(\phi^*)^2]-2\eta_0^\prime E_P[(f^*)^{\prime\prime}(\phi^*)\phi^*\psi]\notag\\
         =& \langle \nabla^2H_{{id}}[\phi^*]\psi, \psi\rangle+\frac{{ \left(E_P[(f^*)^{\prime\prime}(\phi^*)\psi\phi^*]\right)^2}}{E_P[(f^*)^{\prime\prime} (\phi^*)(\phi^*)^2]}\,.\notag
\end{align}
Once again, the magnitude of the second derivative  has been reduced.
 
\subsection{Optimization over affine transformations} Finally, let $\eta_\epsilon$ and $\nu_\epsilon$ be defined by
\begin{align}
    &\sup_{\eta,\nu}\{E_Q[\eta\phi_\epsilon-\nu]-E_P[f^*(\eta\phi_\epsilon-\nu)]=E_Q[\eta_\epsilon\phi_\epsilon-\nu_\epsilon]-E_P[f^*(\eta_\epsilon\phi_\epsilon-\nu_\epsilon)]\,.
\end{align}
{ The existence and smoothness of $\eta_\epsilon$ and $\nu_\epsilon$ are again   guaranteed by convexity and the implicit function theorem.} They satisfy
\begin{align}
    &\partial_\nu|_{\nu=\nu_\epsilon}(E_Q[\eta_\epsilon\phi_\epsilon-\nu]-E_P[f^*(\eta_\epsilon\phi_\epsilon-\nu)])=0\,,\\
    &\partial_\eta|_{\eta=\eta_\epsilon}(E_Q[\eta\phi_\epsilon-\nu_\epsilon]-E_P[f^*(\eta\phi_\epsilon-\nu_\epsilon)])=0\notag
\end{align}
for all $\epsilon$. Simplifying this and differentiating with respect to $\epsilon$ we obtain the two equations
\begin{align}
0=&E_P[(f^*)^{\prime\prime}(\phi^*)\phi^*]\eta_0^\prime-E_P[(f^*)^{\prime\prime}(\phi^*)]\nu_0^\prime+E_P[(f^*)^{\prime\prime}(\phi^*)\psi] \,,\\  
0=&-E_P[(f^*)^{\prime\prime}(\phi^*)(\phi^*)^2]\eta_0^\prime+E_P[(f^*)^{\prime\prime}(\phi^*)\phi^*]\nu_0^\prime\notag-E_P[(f^*)^{\prime\prime}(\phi^*)\phi^*\psi]\,.
\end{align}
If we define
\begin{align}
    &a=E_P[(f^*)^{\prime\prime}(\phi^*)\phi^*]\,,\,\,\, b=E_P[(f^*)^{\prime\prime}(\phi^*)]\,,\\
& c=E_P[(f^*)^{\prime\prime}(\phi^*)(\phi^*)^2]\,,\,\,\,g=E_P[(f^*)^{\prime\prime}(\phi^*)\psi]\,,\notag\\
&h=E_P[(f^*)^{\prime\prime}(\phi^*)\phi^*\psi]\,,\notag
\end{align}
then these have the solution
\begin{equation}
\begin{bmatrix} \eta_0^\prime \\ \nu_0^\prime \end{bmatrix} = 
\frac{1}{bc-a^2} \begin{bmatrix} a & -b \\ c & -a \end{bmatrix}
\begin{bmatrix} g \\ h \end{bmatrix}
\end{equation}
(note that $a^2< bc$ follows from the Cauchy Schwarz inequality together with the assumptions that  $f^\prime$ is strictly increasing and $Q\neq P$). We can compute
\begin{align}
&    \langle \nabla^2H_{affine}[\phi^*]\psi, \psi\rangle\notag\\
=&\frac{d^2}{d\epsilon^2}\big|_{\epsilon=0}\sup_{\eta,\nu\in\mathbb{R}}H[\phi_\epsilon,\eta,\nu]=-E_{P}[(f^*)^{\prime\prime}(\phi^*)(\eta_0^\prime\phi^*+\psi-\nu_0^\prime)^2]\notag\\
    =&\langle \nabla^2H_{{id}}[\phi^*]\psi, \psi\rangle-(\eta_0^\prime)^2c-2\eta_0^\prime h+2\eta_0^\prime\nu_0^\prime a+2\nu_0^\prime g-(\nu_0^\prime)^2b\notag\\
    =&\langle \nabla^2H_{{id}}[\phi^*]\psi, \psi\rangle+\frac{1}{bc-a^2}\begin{bmatrix}
    h& g
    \end{bmatrix}\begin{bmatrix}
b&-a\\
-a&c\end{bmatrix}\begin{bmatrix}
    h\\g
    \end{bmatrix}.\label{eq:eta_nu_gain}
\end{align}
We have $b+c\geq 0$ and $bc-a^2> 0$ hence the matrix in \req{eq:eta_nu_gain} is positive semi-definite.  Therefore the gain term is non-negative and the second derivative is reduced in magnitude, as compared to the unoptimized objective functional \eqref{eq:Hessian_eta1_nu0}. { Finally, we can rewrite the terms $a, c, d$ in \eqref{eq:eta_nu_gain} in terms of $b=E_P[(f^*)^{\prime\prime}(\phi^*)]$ and the tilted measure \eqref{eq:P_star_def}: 
\begin{align}
    &a=E_P[(f^*)^{\prime\prime}(\phi^*)\phi^*]=b\cdot E_{P^*}[\phi^*] \,,\\
 &c=E_P[(f^*)^{\prime\prime}(\phi^*)(\phi^*)^2]=b\cdot E_{P^*}[(\phi^*)^2] \,,\notag\\
    &g=E_P[(f^*)^{\prime\prime}(\phi^*)\psi]=b\cdot E_{P^*}[\psi]\,,\notag\\
&    h=E_P[(f^*)^{\prime\prime}(\phi^*)\phi^*\psi]=b \cdot E_{P^*}[\phi^*\psi]\,. \notag
\end{align}
Then \eqref{eq:eta_nu_gain} becomes
\begin{align}
   & \langle \nabla^2H_{affine}[\phi^*]\psi, \psi\rangle=\langle \nabla^2H_{{id}}[\phi^*]\psi, \psi\rangle+b \cdot \left[ \frac{\cov_{P^*}(\phi^*,\psi)^2}{\Var_{P^*}[\phi^*]}+\left(E_{P^*}[\psi]\right)^2\right]\,.
\end{align}
 
}
\end{IEEEproof}

\section{Exponential Families and the Manifold of Sufficient Statistics}\label{app:submanifold}

Suppose $P=P_{\theta_p}$ and $Q=P_{\theta_q}$ are members of the same exponential family  $dP_\theta=h(x)e^{\kappa(\theta)\cdot T(x)-\beta(\theta)},\,\,\,\theta\in\Theta$  with $T:\Omega\to\mathbb{R}^n$  the vector of sufficient statistics. Then  (under appropriate assumptions) we have 
\begin{align}
&D_f(Q\|P)=\sup_{\phi\in L^1(Q)}\{E_Q[\phi]-E_P[f^*(\phi)]\}=E_Q[\phi^*]-E_P[f^*(\phi^*)]\,, 
\end{align}
and  the explicit optimizer $\phi^*=f^\prime(dQ/dP)$ lies on a (generally nonlinear) $(n+1)$-dimensional manifold of functions, parameterized by the sufficient statistics and constants:
\begin{align}
\phi_{(\beta,\kappa)}=f^\prime(\exp(\kappa\cdot T+\beta)),\,\,\,(\beta,\kappa)\in\mathbb{R}\times\mathbb{R}^n\,.
\end{align}
Therefore, computing the $f$-divergence reduces to the following finite-dimensional optimization problem:
\begin{align}\label{eq:submanifold_Df}
D_f(Q\|P)=&\sup_{(\beta,\kappa)\in\mathbb{R}^{n+1} }\{E_Q[\phi_{(\beta,\kappa)}]-E_P[f^*(\phi_{(\beta,\kappa)})]\}\,.
\end{align}

In some cases one can further reduce the dimension  by optimizing over an appropriate family of transformations.
\begin{enumerate}
  
\item For KL divergence, we have 
\begin{align}
\phi_{(\beta,\kappa)}=\log(\exp(\kappa\cdot T+\beta))+1=\kappa\cdot T+(\beta+1)\,,
\end{align}
 and so the optimizer lies on a linear manifold.  If one optimizes over the family of shifts  (i.e., one uses the Donsker-Varadhan formula) then the $\beta+1$ factor is eliminated and one finds that  
\begin{align}\label{eq:KL_div_manifold}
D_{KL}(Q\|P)=&\sup_{\kappa\in\mathbb{R}^n}\{E_Q[\kappa\cdot T]-\log E_P[e^{\kappa\cdot T}]\}
\end{align}
for any $Q$, $P$ that are members of the exponential family, i.e., the optimization is over the  $n$-dimensional subspace  spanned by the sufficient statistics: $\phi_{\kappa}=\kappa\cdot T,\,\,\,\kappa\in\mathbb{R}^n$. Also, note that $\kappa\to E_Q[\kappa\cdot T]-\log E_P[e^{\kappa\cdot T}]$ is concave.

\item For $\alpha$-divergences, the maximizer of the scaling-improved variational formula (\ref{eq:alpha_div_eta_opt}) lies on the manifold  $\phi_{\kappa}=\exp((\alpha-1)\kappa\cdot T)$, hence
\begin{align}\label{eq:Dfalpha_submanifold}
&D_{f_\alpha}(Q\|P)=\sup_{\kappa\in\mathbb{R}^n}\left\{\frac{1}{\alpha  (\alpha-1)}E_Q[\phi_\kappa]^\alpha E_P[\phi_\kappa^{\alpha/(\alpha-1)}]^{-(\alpha-1)}\right\}\notag-\frac{1}{\alpha(\alpha-1)}\notag
\end{align}
for any members $Q$, $P$, of the exponential family.

\end{enumerate}

{
\section{Asymptotic Variance of the Shift-Optimized $\alpha$-Divergence  Objective Functional}\label{app:asymp_var}

In this section we provide details regarding the computation of asymptotic variance of the objective functional estimators for $\alpha$-divergences (see Theorem \ref{thm:asymptotic_var}), generalizing some of the work on KL divergences from \cite{Song2020Understanding}. The main tool is the following lemma, which is based on the delta method.
\begin{lemma}\label{lemma:delta_method}
 Let $h:\mathbb{R}^d\to\mathbb{R}$ be $C^1$ and $X_n$ be iid $\mathbb{R}^d$-valued  random variables with mean $\mu$, covariance $\Sigma$, and $\sup_n\|X_n\|<\infty$.  Define 
\begin{align}
E_n[X]=\frac{1}{n}\sum_{i=1}^n X_i\,.
\end{align}
 Then
\begin{align}
E[\sqrt{n}(h(E_n[X])-h(\mu))]\to 0
\end{align}
and
\begin{align}
\Var[\sqrt{n}(h(E_n[X])-h(\mu))]\to \nabla h(\mu)\cdot \Sigma\cdot \nabla h(\mu)\,.
\end{align}

\end{lemma}
\begin{IEEEproof}
The central limit theorem  implies
\begin{equation}
\sqrt{n}(E_n[X]- \mu)\xrightarrow{D} N(0,\Sigma)\,.
\end{equation}
The delta method (see, e.g., Theorem 5.15 in \cite{wasserman2013all}) then implies
\begin{align}\label{eq:delta_method}
\sqrt{n}\left(h(E_n[X])-h(\mu)\right)\xrightarrow{D}   N(0,\nabla h(\mu)\cdot\Sigma \cdot\nabla h(\mu))\,.
\end{align}
 Convergence of the mean and variance will  follow from \eqref{eq:delta_method} if we can show uniform integrability of the random variables $|\sqrt{n}(h(E_n[X])-h(\mu))|^2$; see, e.g.,   Theorem 5.9 in \cite{gut2006probability}. The quantities $E_n[X]$ and $\mu$  valued in $B_M(0)$, where $M$ is a bound on $\sup_n\|X_n\|$. The $C^1$ function $h$ is therefore Lipschitz on this ball; let $L$ denote the Lipschitz constant. We can now compute 
\begin{align}
&\sup_n\int 1_{|\sqrt{n}(h(E_n[X])-h(\mu))|^2>c} |\sqrt{n}(h(E_n[X])-h(\mu))|^2dP\\
\leq& L^2\sup_nn\int 1_{nL^2\|E_n[X]-\mu\|^2>c}\|E_n[X]-\mu\|^2dP\notag\\
\leq& L^2\sup_nnE[\|E_n[X]-\mu\|^4]^{1/2} P(nL^2\|E_n[X]-\mu\|^2>c)^{1/2}\notag\\
\leq& L^2\sup_nnE[\|E_n[X]-\mu\|^4]^{1/2}\notag\\
&\times\left( \sum_{k=1}^dP\left(|E_n[X^k]-\mu^k|>\frac{c^{1/2}}{n^{1/2} d^{1/2}L}\right)\right)^{1/2}\,.\notag
\end{align}
 Hoeffding's inequality (see, e.g., Theorem 2.8 in \cite{boucheron2013concentration}) implies
\begin{align}
P\left(|E_n[X^k]-\mu^k|>\frac{c^{1/2}}{n^{1/2} d^{1/2}L}\right)
\leq&2\exp\left(-\frac{n}{2M^2}\left(\frac{c^{1/2}}{n^{1/2} d^{1/2}L}\right)^2\right)\\
=&2\exp\left(-\frac{c}{2dM^2L^2}\right)\,.\notag
\end{align}
 Therefore
\begin{align}\label{eq:unif_int_formula}
&\sup_n\int 1_{|\sqrt{n}(h(E_n[X])-h(\mu))|^2>c} |\sqrt{n}(h(E_n[X])-h(\mu))|^2dP\\
\leq&  2^{1/2}d^{1/2} L^2 \exp\left(-\frac{c}{4dM^2L^2}\right) \sup_nE[n^2\|E_n[X]-\mu\|^4]^{1/2}\,.\notag
\end{align}
If $ \sup_nE[n^2\|E_n[X]-\mu\|^4]<\infty$ then the limit of \eqref{eq:unif_int_formula} as $c\to\infty$ equals zero and we are done. Expanding this expression and using the assumption that the $X_i$'s are iid we find, after a somewhat long but straightforward calculation,
\begin{align}
E[n^2\|E_n[X]-\mu\|^4]
=&\frac{1}{n} E[\|X_{1}-\mu\|^4]+(1-1/n)E[\|X_{1}-\mu\|^2]^2\\
&+ 2\sum_{j,k=1}^d(1-1/n)E[(X_{1}^j-\mu^j)(X_{1}^k-\mu^k)]^2\,,\notag
\end{align}
which is bounded above uniformly in $n$. This completes the proof.
\end{IEEEproof}

We now prove the asymptotic variance result for $\alpha$-divergences, Theorem \ref{thm:asymptotic_var}.
\begin{IEEEproof}
First, we can rewrite the variance as
 \begin{align}
        \alpha^2(\alpha-1)^2n\Var[\widehat{H}_{scale,f_\alpha}[\phi;Q_n,P_n]]
        =&\Var[\sqrt{n} h(E_{Q_n}[\phi],E_{P_n}[\psi_\alpha])]\\
=&\Var[\sqrt{n}( h(E_{Q_n}[\phi],E_{P_n}[\psi_\alpha])-h(\mu))]\,,\notag
    \end{align}
    where
\begin{align}
    h(x,y)\equiv&x^\alpha y^{-(\alpha-1)}-1\,,\,\,\,  \mu\equiv (E_Q[\phi],E_P[\psi_\alpha])\,.
\end{align}
The assumed bounds on $\phi$ imply that $(\phi,\psi_\alpha)$ are valued in a  compact convex set and that $h$ has a $C^1$ extension from this set to all of $\mathbb{R}^2$.  Let $E_n[X]=(E_{Q_n}[\phi],E_{P_n}[\psi_\alpha])$. Note that $E_n[X]$ has mean $\mu$ and covariance $\Sigma=\text{diag}(\Var_Q[\phi],\Var_P[\psi_\alpha])$ (we independently sample from $Q$ and $P$).  The assumptions of Lemma \ref{lemma:delta_method} are now satisfied, and so
\begin{align}
   \alpha^2(\alpha-1)^2n\Var[\widehat{H}_{scale,f_\alpha}[\phi;Q_n,P_n]]       \to \nabla h(\mu)\cdot \Sigma\cdot \nabla h(\mu)\,.
\end{align}
Simplifying this gives the claimed result \eqref{eq:asymp_var_thm}\,.
    \end{IEEEproof}
}

\section*{Acknowledgment}
The research of J.B. and M. K.  was partially supported by the HDR-TRIPODS program of   the National Science Foundation (NSF) under grant CISE-1934846. 
The research of  M.K.  was partially supported by   the Air Force Office of Scientific Research (AFOSR) under  grant FA-9550-18-1-0214. 
Y.P. acknowledges partial support by the project “Innovative Actions in Environmental Research and Development (PErAn)” (MIS 5002358) funded by the Operational Programme ``Competitiveness, Entrepreneurship and Innovation" (NSRF 2014-2020).

\ifCLASSOPTIONcaptionsoff
  \newpage
\fi



%

\bibliographystyle{IEEEtran}
\bibliography{optimizing_variational_representations}

\end{document}